\documentclass[10pt,letterpaper]{article}

\usepackage{ccn}
\usepackage{pslatex}
\usepackage{apacite}
\usepackage{hyperref}
\usepackage{natbib}
\usepackage{lineno}

\usepackage{tabularx}
\usepackage{hyperref}       % hyperlinks
\usepackage{url}            % simple URL typesetting
\usepackage{booktabs}       % professional-quality tables
\usepackage{amsfonts}       % blackboard math symbols
\usepackage{nicefrac}       % compact symbols for 1/2, etc.
\usepackage{microtype}      % microtypography
\usepackage{xcolor}         % colors
\usepackage{tabularx}       % For automatically adjusting table width
\usepackage{caption}        % For better caption handling
\usepackage{adjustbox}
\usepackage{pifont} % For check and cross marks
\usepackage{graphicx}
\usepackage{graphics}
\usepackage{comment}
\usepackage{calc}
\usepackage{float}

\newcommand*{\escape}[1]{\texttt{\textbackslash#1}}
\newcommand*{\escapeI}[1]{\texttt{\expandafter\string\csname #1\endcsname}}

\newcommand{\cmark}{\ding{51}}      % Check mark
\title{The BabyView dataset: High-resolution egocentric videos of infants’ and \protect\\ young children’s everyday experiences}

\author{
\textbf{Bria Long}$^{1,2*}$\footnotetext {*Equal contribution.}  \quad \textbf{Robert Z. Sparks}$^{1*}$ \quad \textbf{Violet Xiang}$^{1*}$ \quad \textbf{Stefan Stojanov}$^{1*}$ \ \quad \textbf{Zi Yin}$^1$ \\ 
\textbf{\quad Grace  E. Keene$^1$ \quad Alvin W. M. Tan}$^1$ \quad \textbf{Steven Y. Feng}$^1$  \quad \textbf{Auddithio Nag}$^1$  \quad \textbf{Chengxu Zhuang}$^3$ \quad \\
\textbf{Virginia A. Marchman}$^1$ \quad \textbf{Daniel L. K. Yamins}$^1$ \quad \textbf{Michael C. Frank}$^1$\\
$^1$Stanford University, 
Stanford, CA 94305 \\
$^2$University of California, San Diego, 
La Jolla, CA 92093 \\
$^3$Massachusetts Institute of Technology, 
Cambridge, MA 02139
\\
\texttt{brlong@ucsd.edu} \\
\texttt{\{bsparks, ziyxiang, stojanov, yinzi, gkeene,} \\
\texttt{tanawm, syfeng, aunag, marchman, yamins, mcfrank\}@stanford.edu}\\
\texttt{chengxuz@mit.edu}\\
}

\begin{document}
% \linenumbers

\maketitle

\section{Abstract}
{
\bf
Human children far exceed modern machine learning algorithms in their sample efficiency, achieving high performance in key domains with much less data than current models. This ``data gap'' is a key challenge both for building intelligent artificial systems and for understanding human development. Egocentric video capturing children's experience -- their ``training data'' -- is a key ingredient for comparison of humans and models and for the development of algorithmic innovations to bridge this gap. Yet there are few such datasets available, and extant data are low-resolution, have limited metadata, and importantly, represent only a small set of children's experiences. Here, we provide the first release of a large developmental egocentric video dataset -- the BabyView dataset -- recorded using a high-resolution camera with a large vertical field-of-view and gyroscope/accelerometer data. This 868 hour dataset includes egocentric videos from children spanning 6 months -- 3 years of age in longitudinal, at-home contexts. We provide gold-standard annotations for the evaluation of speech transcription, speaker diarization, and human pose estimation, and evaluate models in each of these domains.  We train self-supervised language and vision models and evaluate their transfer to out-of-distribution tasks, including syntactic structure learning, object recognition, depth estimation, and image segmentation.  Although performance in each domain scales with dataset size, overall performance is relatively lower than when models are trained on curated datasets, especially in the visual domain. Our dataset stands as an open challenge for robust, human-like AI systems: how can such systems achieve human-levels of success on the same scale and distribution of training data as humans? 
}

\begin{quote}
\small
\textbf{Keywords:} 
head-mounted cameras; data gap; language learning; visual representations; large language models
\end{quote}

Infants and young children are remarkable learners, becoming capable and engaged social partners within their first two years of life. The pace of this developmental progress far exceeds modern machine learning algorithms in its efficiency and capacity \citep{frank2023}. In particular, signature accomplishments of artificial systems, such as few-shot learning \citep{brown2020} and image classification \citep{krizhevsky2012}, require hundreds of billions of words of training data and millions of labeled images. In contrast, human learners become proficient in extending labels for newly learned visual concepts \citep{carey1978acquiring} and producing language \citep{frank2021variability} from only tens of millions of words and far fewer labeled examples \citep{zhuang2021}.
This ``data gap'' between human and machine learners is thus a key challenge for the joint goals of understanding human learning and building intelligent artificial systems. Making progress will require not just an understanding of the flexibility of human intelligence, but also an understanding of the efficiency of human learning. 

Data availability is a major barrier to progress in our understanding of the gap in learning efficiency between machines and humans. 
To make effective comparisons between human and machine learners, we need to be able to evaluate models on data comparable to what children see and hear during everyday learning experiences. 
While models today are trained on millions or billions of images and/or videos, these are taken from the adult perspective, providing a very different vantage point on the world that is disconnected from real-world learning environments. 

Egocentric video recordings taken from the child's perspective provide a key window into what children both see and hear as they learn about the world around them and from their social partners \citep{smith2015,yoshida2008,aslin2009infants,franchak2011}. Developmental psychology studies using these types of video recordings have together revealed that the infant view is dramatically different from that of adults' \citep{yoshida2008} and varies as children learn to locomote on their own and interact actively with the objects, places, and people around them \citep{kretch2014,long2022a}. 

Here we present the largest high-resolution developmental egocentric video dataset to date, the BabyView dataset. We collect videos from 31 families predominantly located in the United States, totaling 868 hours of usable recordings. We capitalize on innovations in the development of head-mounted cameras \citep{long2023}, obtaining videos with a large vertical field of view and coordinated gyroscope/accelerometer data that can be used to estimate the child's own head movements. We provide pose detection, automated speech transcriptions, and diarization, along with gold-standard annotations for use in evaluating each of these. 
We additionally provide language outcome measures for a subset of the children in the dataset. 
We then evaluate self-supervised vision and language models on these data relative to existing benchmarks. 
 
\section{Related Work}

\paragraph{Few developmental egocentric video datasets are available} Egocentric video has been an important domain for computer vision \citep{damen2022, grauman2022} and resulting commercial applications, such as wearable devices.
Yet, egocentric video datasets are mostly taken from the adult perspective, including the Ego4D dataset, which has become an important standard in this field \citep{grauman2022}. 
Head-mounted cameras have also been used in research with children, including both descriptive investigations \citep{yoshida2008,aslin2009infants,franchak2011,kretch2014,fausey2016faces, bergelson2017} and computer vision studies \citep{sheybani2024curriculum, zhuang2021}. 
Unfortunately, most prior work did not obtain consent for broad sharing with other research groups and so many major datasets are unavailable for re-analysis. 

The few developmental egocentric video datasets that are available have been difficult to use for training models for reasons of both data quantity and quality \citep{long2022a,sullivan2021,bergelson2017}. For example, the SAYCam dataset -- by far the largest available dataset -- is relatively low-resolution (480 $\times$ 640 pixels), has limited motion-correction (leading to blurry views), and has timestamps imprinted on every frame \citep{sullivan2021}. The audio quality is quite variable depending on the background noise and context, and the videos have restricted vertical view angle that obscures views of children's hands and what children are interacting with. Further, SAYCam represents video from three children of highly-involved and informed academic parents, all of whom were the first children in their families. These issues have limited the field's ability to make use of automated annotations of the visual or linguistic content of these videos and have restricted the ability to use these data to draw broadly generalizable conclusions. Here, we present the largest high-resolution, developmental egocentric video dataset with broad consent from caregivers for reuse within the research community.

\paragraph{Models trained on developmental data show limited performance} Self-supervised vision models trained using developmental egocentric video data \citep{zhuang2021, orhan2020self, zhuang2022well, orhan2024learning, vong2024} have had some intermediate success. However, these models significantly underperform those self-supervised models trained on curated datasets, while the latter models approach the accuracy of models trained using fully-supervised methods \citep{oquab2023dinov2, dinov1, mae, simclr, moco}.
Thus, it remains unclear whether the current state-of-the-art techniques represent truly general-purpose visual learning algorithms.
In particular, it is unclear whether gaps in model performance are due to dataset quality and quantity or are instead due to the difficulty of learning robust representations from children's more realistic everyday inputs.

Relatedly, in the language domain, recent work has investigated the possibility of training language models (LMs) on small-scale developmental datasets \citep[see e.g.,][]{warstadt2023findings, zhuang2024lexicon, feng2024childdirectedspeecheffectivetraining}, but most of these have focused on datasets larger than those available from egocentric video data. For example, the text data used in the popular BabyLM competition \citep{warstadt2023findings} are also meant to approximate what a 10-year-old child could receive (including text from Wikipedia and other sources), which is very likely more -- and different -- data than what is required to acquire a language. One exception is \citet{qin2024systematic}, who trained GPT-2 \citep{radford2019language} on very small amounts of input from a single child and investigated the amount of grammatical knowledge that could be learned. 

Here, we evaluate whether data from a new, high-resolution dataset will lead to increases in performance for self-supervised visual and linguistic benchmark models.

\section{The BabyView Dataset}

We address gaps in data availability by collecting and analyzing a new set of developmental egocentric videos: the BabyView dataset. The current paper describes the first release of the dataset, but data collection is still ongoing and we anticipate future growth in the overall size of the dataset.
Recordings were obtained using a high-resolution head-mounted camera for infants and children from 6 months through 3 years of age in at-home settings.
In the BabyView-Home portion of the dataset, 31 families recorded longitudinal data during everyday activities for a total of 868 hours across all children.
All videos are accompanied by accelerometer/gyroscope data that can be used to estimate children's head-motion \citep{joshi2010image, karpenko2011digital,joshi_gopro_icra_2022}. We additionally release the Ego-SingleChild dataset, a related dataset with 70 hours of recordings with a different camera (see below).
Together, these data comprise the first release of the largest high-resolution egocentric video dataset from children that will be available to researchers for both descriptive analysis and model building (see Table \ref{tab:dataset_info} for comparison to prior datasets).

\begin{table*}[t!]
    \caption{Comparison of the BabyView dataset to existing related datasets; the BabyView dataset is the only egocentric developmental video dataset with accelerometer/gyroscope data available for research. }
\scalebox{0.9}{
    \label{tab:dataset_info}
        \begin{tabularx}{\textwidth+1.5cm}{l p{1.6cm} p{1.5cm} p{1cm} p{1cm} p{1cm} p{1cm} p{1.5cm} p{.5cm}}
        \toprule
        Dataset & Egocentric? & Longitudinal? & Type & N & Hours & Audio & Transcript & Motion \\
        \midrule
        BV-Home &  \cmark  &  \cmark  &  Infant  &  31  &  868  &  \cmark  &  \cmark  &  \cmark \\
        % BV-Preschool &  \cmark &   &  Child  &  39  &  63  &  \cmark  &  \cmark  & \cmark \\
        Ego-SingleChild &  \cmark  &  \cmark  &  Infant  & 1 &  70  &  \cmark  &  \cmark  &   \\
        \midrule
        SAYCam~\cite{sullivan2021} & \cmark & \cmark & Infant &  3  &  476  &  \cmark  &  \cmark  &    \\
        Ego4D~\cite{grauman2022} & \cmark &   & Adult & 931 & 3,670 & \cmark & \cmark &  \\
        Epic Kitchens~\cite{damen2018scaling} & \cmark &    &  Adult  &  37  &  100  &  \cmark  &  \cmark &    \\
        \bottomrule
    \end{tabularx}    
    }
    % \vspace{-8pt}
\end{table*}

\subsection{Camera and sensor data}

\label{sec:camera}

\begin{figure*}[t]
  \centering
  \includegraphics[width=\linewidth]{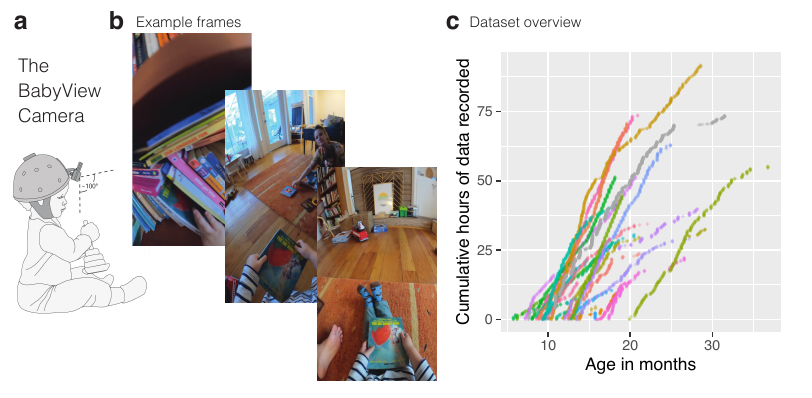}
  \vspace{-23pt}
  \caption{(a) Schematic of a child wearing the BabyView camera illustrating a large vertical field of view.
  (b) Example frames from a video in the dataset (where the parent has provided broad sharing consent). (c) Cumulative hours of video by each of the participants; each color represents an individual child. The grey line represents video data from the Ego-SingleChild subset. Data collection is ongoing. The BabyView dataset thus collects high-resolution video and gyroscope/accelerometer with a large vertical field of view from many children over a large age range, with dense longitudinal data from a few participants.}
  \label{fig:dataset-overview}
  % \vspace{-.5cm}
\end{figure*}

The BabyView camera is a GoPro Hero Bones camera attached to a child-safety helmet.
This camera was selected because it has gyroscope and accelerometer data, built-in image stabilization features, and relatively high-resolution sound and video \citep{long2023}.
The camera is oriented vertically and is neutral with respect to the face plane of the child, enabling the camera to capture both adult faces and objects in a child’s hands in the same image, with an effective view angle of 100° vertical by 75° horizontal (see Figure ~\hyperref[fig:dataset-overview]{1a,b}) \citep{long2023}.
% \footnote{A brief overview of the camera used for data collection can be seen at \href{https://langcog.github.io/babyview/.}{https://langcog.github.io/babyview/.}}  

\subsection{Dataset components}

\paragraph{BV-Home} 
Thirty-one families consented to capture home recordings with their infant-toddler (0;5--3;1 years, average age at onboarding = 10 months, SD = 0.26 years, see Figure~\hyperref[fig:scaling-exp]{1c}). Data collection is ongoing. Families were recruited from a convenience sample of researchers in the field of cognitive development (N=7/31 families) and from advertisements within the State of California, and the broader United States. Some English-speaking (N = 18/31) and English/Spanish bilingual families (N=1/31) completed one or more parent-report measures of children’s language development using the long-forms of the MacArthur-Bates Communicative Development Inventories \citep{marchman2023, jacksonmaldonado2003}. Our current sample is relatively multilingual (with only 19/31 English monolingual participants) and
highly educated, with 24/31 families having at least one parent with a graduate degree.  See Appendix for further information on participant consent, detailed demographics, and number of language questionnaires.

\paragraph{Ego-SingleChild}
We also release 70 hours of data from a single child of an academic parent who recorded frequently. \textcolor{black}{This participant was recruited before procedures were finalized, and recorded using a different camera from other participants. They used a Cigno F18 Night Vision 1080P Headband Sport Camera rather than the BabyView camera, which yields shorter and lower-resolution videos. However, they are comparable to previous dense longitudinal recordings (see \citep{sullivan2021}), and thus provide additional data that other researchers may benefit from.} 
 
\subsection{Data access \& ongoing data collection}

Egocentric video data from children in their home environments necessarily contain more sensitive information than videos in egocentric videos by adults. Families provided full consent for the data that are shared at the time of recording. During a 6-month period after recording, families can also retract any portion of their recording. Thus, all data in this release will be made available in August 2025 once the parental embargo period has lapsed for all videos in this release  (release 2025.1).  To ensure BabyView data are accessible to researchers while protecting the privacy of participants, we distribute the data through Databrary (\href{https://nyu.databrary.org/}{https://nyu.databrary.org/}) \citep{gilmore2016curating}, similar to previous developmental egocentric datasets \citep{sullivan2021,bergelson2017}. Databrary is a U.S. National Institutes of Health-funded site designed specifically for the distribution of developmental video data. Access to data on Databrary requires investigators to be authorized via an institutional agreement that bars re-identification of participants and redistribution of data. 

BabyView is an ongoing longitudinal project and our aim is to release further data as the dataset grows. Because of the multi-faceted and growing nature of our dataset, we do not pre-specify train/test splits, recognizing that any split might be appropriate for only a subset of research goals (e.g., examining age-related change, or within- vs. cross-child change).

\begin{table*}[t!]
    \caption{Performance of the automated transcription and diarization pipeline across the age of the child and the speaker. Child-produced speech had the highest error rates. }
    \centering
\label{tab:language_annotations}
        \begin{tabularx}{\textwidth}{XXXXXXX}
        \toprule
        Dataset & Child age & Speaker & Word error rate & Diarization \newline precision & Diarization \newline recall & Utterances \newline annotated \\
        \midrule
        BV-Home & All Ages & All Speakers & 0.35 & 0.66 & 0.66 & 2242 \\
         &  6-18 m.o. & Adult &  0.26  & 0.66 & 0.79 & 1371 \\
         &   & Key-child &  1.08  & 0.73 & 0.45 & 208 \\
         &   & Other-child &  0.51  & 0.63 & 0.39 & 95 \\
         &  18-30 m.o. & Adult &  0.37  & 0.64 & 0.77 & 271 \\
         &   & Key-child &  0.56  & 0.76 & 0.62 & 94 \\
         &   & Other-child &  0.21  & 0.60 & 0.38 & 15 \\
        \bottomrule
        \end{tabularx}
\end{table*}

\section{Annotations}

\subsection{Language annotations}

\paragraph{Transcription \& diarization pipeline}
All videos were transcribed using Distil-Whisper model distil-large-v3 \citep{gandhi2023distil}. As this version only supports English transcription, we conducted transcription validation for families who reported speaking only English at home.  We also ran a multilingual voice type classifier \citep{lavechin2020open} in parallel on the audio extracted from all BabyView-Home videos (regardless of language), which classified the speech segments as originating from a female adult, male adult, key child (the wearer of the camera), or other child. \textcolor{black}{Transcripts and diarizations were then merged: Each utterance was assigned to one speaker by choosing the model-annotated speaker category that had the greatest overlap with the utterance timestamps. In some cases, an utterance did not overlap with any model-annotated speaker; these were marked as NA (NA rate was 8.39\% for BV-Home).} For our language model training experiments below, we also ran the same pipeline on the SAYCam audio, though we did not conduct validation on this dataset. 

\paragraph{Evaluation procedure}
To assess the accuracy of speech transcription and speaker diarization on this dataset, 
we hand-annotated a subset of 2242 utterances in English monolingual families, stratified across participant and age at the time of recording. These utterances account for 2.59 hours of the BV-Home videos.
% Each session was divided into video segments of approximately 12 minutes, as a result of the internal caching mechanism of the GoPro camera. 
For each sampled video, we extracted 30 seconds of video beginning at the midpoint of the video. 
Two authors transcribed the speech and labeled the speaker in each segment.
For transcription validation, we computed a Word Error Rate (WER), which is the ratio of the number of word-level errors to the total number of words in the original utterance \citep{gandhi2023distil}. To evaluate speaker diarization accuracy, we computed precision and recall of the model output by age and speaker. 

\paragraph{Child-produced and child-directed speech is challenging for transcription algorithms}
Across all speakers, WER for automated transcriptions was comparable to that for adult recordings and was somewhat lower in these naturalistic home environments, especially compared to that previously seen using the same methods with preschool classroom recordings \citep[see][]{sparks2024preschool}. Qualitatively, these decrements in performance appear to result from a high prevalence of infant-directed speech with which annotation algorithms are less familiar.
Although automated transcriptions perform poorly for the youngest children, we see considerable improvement in WER of child-produced speech of the older (18--30 months) children in the dataset.
Similarly, we found that Whisper often hallucinated incorrect utterances for child-produced speech for the youngest infants (rather than appropriately labeling it as babble).
The speaker diarization algorithm \citep{lavechin2020open} was able to identify whether a child vs. adult was speaking 78\% of the time, and often could accurately identify the speaker type (female-adult, male-adult, key-child, other-child) 
 (see Table \ref{tab:language_annotations}).
While combining speaker diarization and automated transcriptions can be useful, 
modern transcription algorithms are still  less accurate than humans at understanding both adult speech in home environments and child-produced speech.

\subsection{Human pose annotations}

\paragraph{Pose annotations}  Human pose annotations provide critical data on the social information available to infants and children throughout development that could guide their learning \citep{fausey2016faces, long2022a} and pose detection models been successfully applied to previous egocentric datasets \citep{long2022a, long2022b}. We evaluated how well state-of-the-art pose detectors performed on the BabyView dataset. To do so, we first sampled 353 frames from the dataset (stratified across participants and sessions) and manually annotated the 333 non-blurry frames using LabelStudio \citep{LabelStudio}, creating a validation set. To efficiently annotate the frames, we deployed the RTMPose \citep{jiang2023rtmpose} model via MMPose \citep{mmpose2020} as a backend to provide initial pose keypoints and bounding box predictions, which we then manually corrected. The pose annotations followed the format used in the COCO keypoints dataset \citep{lin2014microsoft, sun2019deep}. To evaluate the accuracy of keypoint detections and compare our results with those of other studies, we adopted the Object Keypoint Similarity (OKS) metric \citep{sun2019deep} (see Appendix).

\begin{table*}[t]
\caption{Pose Detection performance on COCO2017 Val and BabyView Val. BabyView Validation frames were more challenging than COCO for all models except ViTPose-H. AP refers to average precision, and AR refers to average recall.}
\centering
\scalebox{.95}{
\begin{tabular}{lcccccc}
\toprule
Architecture & Num. Params & Input Size & COCO-AP & Babyview-AP & COCO-AR & Babyview-AR\\
%& & & AP & AP & AR & AR \\
\midrule
RTMO-l \citep{lu2023rtmo} & 44.8M & 640x640 & 0.724 & 0.593 & 0.762 & 0.723 \\
YOLOXPose-l \citep{maji2022yolo} & 87.0M & 640x640 & 0.712 & 0.588 & 0.749 & 0.658 \\
SIMCC-resnet50 \citep{li2022simcc} & 25.7M & 384x288 & 0.735 & 0.676 & 0.790 & 0.723 \\
RTMPose-l-aic-coco \citep{jiang2023rtmpose} & 36.7M & 384x288 & 0.773 & 0.735 & 0.819 & 0.773 \\
HRFormer-pose-base \citep{NEURIPS2021_3bbfdde8} & 43.2M & 384x288 & 0.774 & 0.743 & 0.823 & 0.785 \\
ViTPose-H \citep{xu2022vitpose} & 632M & 256x192 & 0.788 & 0.788 & 0.840 & 0.825 \\
\bottomrule
\end{tabular}
}
\label{tab:pos_results}
\end{table*}

\paragraph{Child egocentric viewpoints are challenging for most pose detection models} The BabyView validation set was more challenging for most models than the COCO validation set \citep{lin2014microsoft}, highlighting a new pose benchmark for naturalistic egocentric videos (see Table \ref{tab:pos_results}). However, ViTPose-H, the largest model in the group, showed comparable performance between the two validation sets, suggesting that it is more robust to the viewpoint variation inherent in egocentric videos.

\section{Benchmarks}

\subsection{Language representation learning}
Next, inspired by the BabyLM challenge, which seeks to learn human-like linguistic representations from small amounts of developmentally-realistic data \citep{warstadt2023findings}, we examined the ability to learn linguistic representations from the BV-Home transcripts. We compared our results with those obtained using high-quality data from the Child Language Data Exchange System (CHILDES), a repository of human-transcribed corpora of children's and caregivers' talk \citep{macwhinney2014childes}. 

\paragraph{Experiment Setup} We pretrained GPT-2 \citep{radford2019language} with 124M parameters (small) on each dataset for up to 20 epochs (see Appendix for details). We trained three seeds for each model, averaging their evaluation results. For BV-Home, after deduplication, we only used transcripts for English monolinguals for a total of $\sim$1.3M words of conversation (19 families, 469 hours of video data), corresponding to $\sim$2M total words including speaker labels and other metadata. To match this, we sampled %$\sim$1.3M words of conversation (out of $\sim$2M words) from SAYCam, also totaling $\sim$2M total words.%The SAYCam transcripts were not diarized.
2M total words from the diarized SAYCam data. For contrast, the total amount of human-transcribed English-language data available in CHILDES (including speaker labels and other metadata) is $\sim$29M total words. To align the amount of training data across datasets, we sampled 2M total words from CHILDES.

We further compared training on the combination of BV-Home and SAYCam data with 4M total words from CHILDES. We also trained a version on the entire 29M English subset of CHILDES, in line with \citet{feng2024childdirectedspeecheffectivetraining}. Each dataset was separated into 85/15 train and validation splits. This resulted in 1.7M/0.3M splits for the 2M experiments, 3.4M/0.6M splits for the 4M experiments, and 24.5M/4.5M splits for the 29M experiment. For evaluation, we used Zorro \citep{huebner-etal-2021-babyberta}, a benchmark compatible with child vocabulary that aims to quantify the grammatical knowledge of LMs by assessing their capability to effectively distinguish between minimal pairs of sentences that exhibit various grammatical contrasts. 

\begin{figure}[t]
  \centering
\includegraphics[width=\linewidth]{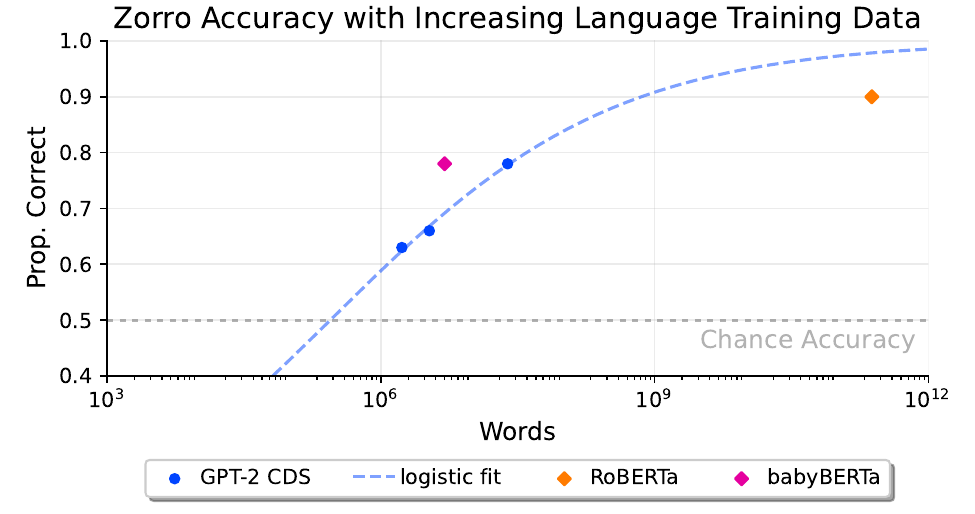}
  \vspace{-5pt}
  \caption{Language data scaling experiments, showing grammatical accuracy on Zorro (chance = 0.5) for GPT-2 trained on progressively increasing amounts of child-directed speech (CDS) data. Within the GPT-2 CDS data points, the first represents 1.7M training words from BV-Home, the second represents 3.4M training words combined over BV-Home and SAYCam, and the final point represents 24.5M training words from CHILDES. Zorro accuracy is also shown for RoBERTa \citep{liu2019robertarobustlyoptimizedbert} [240M words] and BabyBERTa \citep{huebner-etal-2021-babyberta} [5M words]. Given the clear saturation of the metric in the larger pretrained model, we used a logistic function, which asymptotically approaches 1, rather than a linear fit.} %}
  \label{fig:scaling-exp-lang}
  \vspace{-10pt}
\end{figure}

\begin{figure*}[t]
  \centering
  \includegraphics[width=\linewidth]{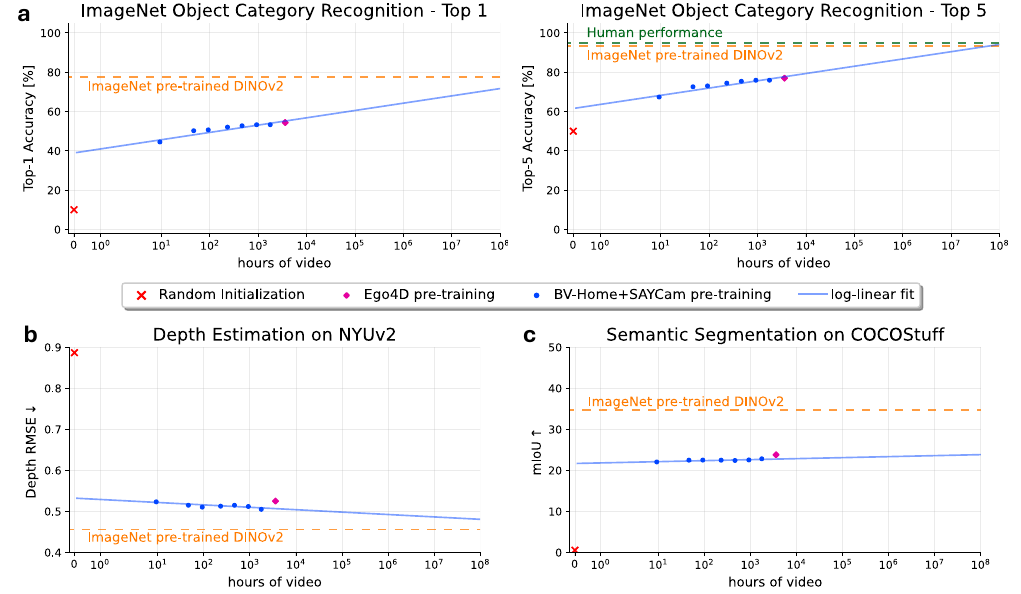}
  \vspace{-15pt}
  \caption{Data scaling experiments for object recognition, depth estimation and semantic segmentation. In \textbf{a} we observed a trend that DINOv2 would require upwards of $10^7$ hours of video to match human or ImageNet self-supervised ImageNet performance. In \textbf{b} and \textbf{c}, we also observed unfavorable scaling for depth estimation and semantic segmentation.}
  \label{fig:scaling-exp}
%\vspace{-20pt}
\end{figure*}

\paragraph{BV-Home transcriptions provide comparable learning signal for grammatical knowledge} All GPT-2 models achieved above-chance performance on the Zorro evaluation, even with fewer than 2M words of total training data. %(see Appendix for complete results).
For the 2M experiments (1.7M training words), there was a negligible difference between BV-Home (63.47$\pm$1.24\%), SAYCam data (63.34$\pm$1.99\%), and CHILDES (63.70$\pm$1.17\%). For the 4M experiments (3.4M training words), CHILDES (66.59$\pm$1.63\%) performed slightly better than the combination of BV-Home and SAYCam (65.13$\pm$1.33\%). Training on the full CHILDES (24.5M training words) resulted in significantly higher performance (78.29$\pm$0.51\%), as expected with much more language data. This is also shown in Figure \ref{fig:scaling-exp-lang}; training on more language data resulted in better performance, in contrast to our vision data scaling experiments shown in Figure \ref{fig:scaling-exp}. Overall, despite the potential data quality issues in BabyView and SAYCam transcripts (introduced by speech recognition errors), we observed that transcriptions of BV-Home and SAYCam are overall comparable to CHILDES as a learning signal for language models to obtain grammatical knowledge.

\subsection{Visual representation learning}

We conducted a first set of experiments to investigate the ability of recent self-supervised models to learn useful visual representations from frames taken from these egocentric videos. Enabled by BV-Home, we conducted the largest scale evaluation to date of self-supervised learning methods trained on children's egocentric visual experience.

\textbf{Experiment Setup} We trained a ViT-B/14 DINOv2~\citep{oquab2023dinov2} from scratch as our reference self-supervised learning algorithm, due to its high performance on various downstream tasks, including object recognition, depth estimation, and semantic segmentation. We used the standard training configuration from the official code base across all training runs. We sampled Ego4D at 1 FPS, leading to 15M frames, and sampled the BV-Home and SAYCam at 5FPS, leading to about 8M frames per dataset for initial comparisons, and another 8M frames from the full BV-Home dataset for a total of 16M BV-Home frames. Despite the inherent redundancy in video data, this ensured a relatively large amount of data, compared with the 1.4M ImageNet training set. We evaluated object recognition accuracy on ImageNet, and after additional training on high-resolution images of the original datasets, we evaluated depth estimation on NYUv2~\citep{silberman2012indoor} and semantic segmentation on COCOStuff~\citep{caesar2018coco}. On top of the frozen ViT, for ImageNet we used kNN and a linear probe, whereas for depth estimation we trained a DPT and for semantic segmentation we used a linear probe, following  DINOv2 protocols (see Appendix).

\begin{table*}[t!]
\centering
\caption{Object recognition, depth estimation, and semantic segmentation results on the BabyView \& comparison datasets. Downstream generalization accuracy was significantly reduced when learning on frames from egocentric videos relative to curated datasets.}

\scalebox{1.0}{
\begin{tabular}{lccccc}
\toprule
& \multicolumn{2}{c}{Object Recognition -- Top 1} & Depth Estimation & Semantic Segmentation \\
\textbf{Dataset} & ImageNet kNN$\uparrow$ &  ImageNet linear$\uparrow$ & NYUv2 RMSE$\downarrow$ & COCOStuff mIoU$\uparrow$ \\
\midrule
None (random init.)                     & 10.00  & 1.43 & 0.886 &  0.54  \\
\midrule
LVD-124M~\citep{oquab2023dinov2}         & 82.10 & 84.50 & 0.307 & 44.46  \\
ImageNet~\citep{russakovsky2015imagenet} & 76.29 & 77.64 & 0.456 & 34.65  \\
\midrule
Ego4D\citep{grauman2022}                                  & 43.59 & 54.39 & 0.525 & 23.78      \\
SAYCam\citep{sullivan2021}                                  & 42.59 & 52.52 & 0.518 & 21.08  \\
BV-Home (430 hours)                                & 40.72 & 52.19 & 0.526 & 22.03  \\
SAYCam + BV-Home (430 hours)                        & 41.76 & 53.28 & 0.511 & 22.53  \\
SAYCam + BV-Home (868 hours)                        & 42.76 & 53.28 & 0.505 & 22.81 \\
\bottomrule
\end{tabular}
}
\label{tab:dataset-comparison}
%\vspace{-15pt}
\end{table*}

\paragraph{Self-supervised learning from any egocentric data is challenging} We anticipated that the more diverse and higher-resolution videos in BV-Home would afford improvements over prior egocentric video datasets \citep{sullivan2021}. Yet, we found that models trained on a similar amount of BV-Home data did not outperform those trained on the SAYCam dataset, despite the difference in data quality (see Table~\ref{tab:dataset-comparison}), though we found a small improvement in semantic segmentation performance on models trained on BV-Home vs. SAYCam.\footnote{ Note results are above random chance: ImageNet -- 0.001, NYUv2 -- 2, COCOstuff -- 0.2.} More broadly, however, we found that the gap in performance is not just specific to data collected from children. Even when training on Ego4D -- a roughly 7$\times$ larger and more diverse dataset -- we saw that a significant gap to curated vision datasets remained across all tasks. We further investigated training an additional self-supervised learning method, MoCov3~\citep{chen2021empirical} also based on a ViT-B/16 on 430 hours of BV-Home data: We obtained 18.7 for kNN and 27.3 for linear on ImageNet, indicating that other self-supervised learning techniques also show a significant gap in performance.

\paragraph{Insufficient scaling to meet human or self-supervised performance from curated datasets}
Given a reasonably large amount of training data from egocentric video of children's visual experience, could the current self-supervised state-of-the-art model reach human performance, or obtain equivalent performance to training on curated vision datasets? 
To address this question, we trained on 1\%, 5\%, 10\%, 25\%, 50\%, and 100\% of a combined dataset of our first half of BV-Home (430 hours) and SAYCam, and extrapolated by fitting log-linear trend lines. For a final datapoint, we added another 424 hours of BV-Home video data (see Table 4). For object recognition on ImageNet (see 
Figure ~\hyperref[fig:scaling-exp]{3a}) we observed that more than $10^7$ hours would be required to reach human performance~\citep{russakovsky2015imagenet} or ImageNet pre-training performance. 
In Figures~\hyperref[fig:scaling-exp]{3b} and~\hyperref[fig:scaling-exp]{3c}, we found that a similar trend holds for depth estimation and semantic segmentation, with saturating performance as the scale of data is increased, even as we roughly doubled the amount of BV-Home data used in pre-training.
Note that the first two points on these plots indicated 160K and 800K images and the last point 16M images.
While training on the combination of both SAYCam and the largest draw of the BV-Home dataset did lead to a numeric improvement evaluation metrics compared to SAYCam alone, the improvement was relatively slim.
Overall, these results suggested that there is still a substantial ``data gap" between state-of-the-art self-supervised vision models and children.

\section{General Discussion}
We present a new, large-scale, high-resolution egocentric video dataset documenting infants' and young children's everyday experiences, accompanied by both dense metadata and gold-standard annotations for several key domains. In contrast to prior work with lower-resolution videos and earlier models \citep{long2022b}, we find that state-of-the-art speech recognition \citep{gandhi2023distil,radford2023robust} and pose detection \citep{xu2022vitpose,mmpose2020} models perform well on stratified samples of frames and audio recordings from the dataset. Further, language models trained on these data performed comparably to models trained on current gold-standard corpora of hand-transcribed child-directed speech corpora.
The BabyView camera thus provides improved data over which supervised algorithms can extract descriptives that will be an important resource for characterizing children's early learning environments \citep{sparks2024preschool}. 

Yet, our results also suggest that the naturalistic, everyday experiences of children pose a challenging problem for the most advanced learning algorithms, especially in the visual domain: current state-of-the-art models fall short relative to existing benchmarks when trained on ``human amounts'' of visual or linguistic data, requiring unrealistic amounts of additional data to achieve human-level performance \citep{frank2023}. Scaling for language data suggested the possibility of relatively strong performance with human-scale data, but scaling for vision models was strikingly bad.  In particular, our results suggest that current self-supervised visual learning models are dependent on large, curated datasets with a broad diversity of inputs to construct robust visual representations useful for object recognition, depth estimation, and semantic segmentation. 

While a similar ``data gap'' has also been reported by \citet{orhan2021much}, our findings yield a somewhat lower estimation of the amount of data needed to achieve human-level performance than suggested in \citet{orhan2021much} and higher estimation than in \citet{orhan2023scaling}.  However, direct comparisons are challenging due to the fact that our results use the state of the art DINOv2~ \citep{oquab2023dinov2, orhan2021much} rather than masked autoencoders ~\citep{mae,orhan2023scaling}, and solely egocentric video datasets (vs. a mix of egocentric and allocentric). Most importantly, we do not fine-tune our models end-to-end on ImageNet, as in \citet{orhan2021much, orhan2023scaling}, but freeze the pre-trained encoders and train readout layers on top. Overall, however, our results are broadly convergent with \citet{orhan2021much} in suggesting that there remains a large data gap between human and machine performance in self-supervised visual learning. 

 \textcolor{black}{What accounts for this difference between the visual and linguistic domains? We suspect that this is because the language data in these experiments are closer to the standard data used to train large language models---e.g., conversation transcripts or subtitles. In contrast, images sampled from egocentric videos vary dramatically from images in curated visual datasets. Using transcribed language also means that the language information has undergone some segmentation and parsing, unlike the visual information. Future work that systematically varies input data will help confirm these ideas.}

What might lead to more child-like models of early learning?
One idea is that the joint learning of visual and language representations \citep{vong2024} requires more fine-grained and efficient learning algorithms, such as lexicon-level visual grounding~\citep{zhuang2023visual, zhuang2024lexicon}.
In addition, children's everyday experiences contain deep regularities within everyday activity contexts \citep{clerkin2017real, clerkin2022real, de2022ten} that are challenging for current models but appear advantageous for human learners.
For example, the same objects and words are repeated often within the same activity contexts (e.g., mealtime), which could create known contexts for children to learn infrequent items (e.g., pomegranates).
Constructing models that can learn as children do from these skewed input distributions -- where some words and objects are frequent and others appear very rarely -- is thus a key challenge for future work \citep{smith2017developmental}.

In addition, we speculate that focusing on modeling event representations in naturalistic video \citep{zhuang2020unsupervised} might lead to more developmentally realistic models, as children are, of course, learning from continuous events rather than static images. Indeed, new work suggests that video models trained on SAYCam \citep{sullivan2021} can learn action concepts \citep{orhan2024self}. In addition, we suspect that incorporating information about both children's own head-motion  \citep{joshi_gopro_icra_2022} via IMU data (contained in this dataset), as well as attentional guidance signals from caregivers \citep{long2022a, yu2021infant} may yield more data-efficient models of early language and visual concept learning.  

Regardless of innovations in model architectures or learning algorithms, our results highlight the need for developmentally-appropriate outcome data \citep{tan2025devbench} which can be used to evaluate models trained on developmental data. Our speculation about scaling for language model training is hindered by the lack of developmental data on grammaticality judgments -- we do not know what a realistic topline should be for human learners. Similarly, toddlers cannot classify all ImageNet categories, and a growing literature suggests that object recognition abilities mature throughout middle childhood, as does their visual concept knowledge \citep{long2024parallel, huber2023developmental}. Children's emerging mid-level visual understanding (such as motion, 3D shape, and depth perception) may also be an alternative basis for comparing models and children, especially as children may develop these capacities through active exploration and interaction with their world long before they have fine-grained category representations.  \textcolor{black}{Thus, this work highlights the need to create benchmarks that allow us to measure and quantify performance in a unified way across both models and children. Systematically comparing models' and children's emerging representations across both the linguistic (e.g., lexical knowledge, grammar, semantics) and the visual (e.g., visual concepts, intuitive physics) domains will likely help elucidate the observed gap in model performance by yielding testable predictions for future work. } 

 \textcolor{black}{Finally, models trained on the same visual diet as children might eventually emerge as superior models of neural responses to visual stimuli. While there is a fairly substantial literature predicting neural responses in ventral cortex from the embeddings of models trained in different ways \citep{yamins2014performance, conwell2024large}, very few of these models have been trained on naturalistic datasets (cf. \citep{zhuang2021, conwell2024large}). We anticipate that the BabyView dataset will enable the systematic benchmarking of models trained with more ecologically-valid input data.} 

These data have several limitations. First, these data necessarily incorporate selection bias: parents who opt-in to the study are recording in their homes when they choose to (to avoid privacy issues) and can choose to excise any portion of their data; in addition, some naturalistic experiences (e.g., bathtime) are not incorporated into the dataset.
Further, with two exceptions, all families are located in the United States, limiting generalizability. 
Nonetheless, BV-Home incorporates data from a greater diversity of families across race, ethnicity, and family incomes than before (see Appendix). 
The potential harms that could arise from this dataset relate to breaches of privacy and trust on the part of the participating families. 
To guard against these, researchers are required to sign the Databrary data use agreement \citep{gilmore2016curating}, which prohibits reidentification or redistribution of videos.

In sum, we present the first release of a new, large-scale, high-resolution developmental egocentric video dataset. Our dataset provides an unprecedented view into the everyday experiences of young children and stands as a challenge to modern AI: how can such systems achieve human levels of success on the same scale and distribution of training data as human children?

\section{Acknowledgments}
We gratefully acknowledge the families who contributed to the dataset. This work was supported by an NIH R00HD108386 to B.L., a Schmidt Futures gift to M.C.F., a Stanford Human-Centered Artificial Intelligence Institute Hoffman-Yee Grant to M.C.F. and D.L.K.Y, a gift from the Center for the Study of Language and Information Olney Fund, and a gift from Amazon, Inc.. D.L.K.Y. and trainees were supported by a Simons Foundation grant 543061, National Science Foundation CAREER grant 1844724, National Science Foundation Grant NCS-FR 2123963, Office of Naval Research grant S5122, ONR MURI 00010802, ONR MURI S5847, and ONR MURI 1141386 - 493027. We also thank the Stanford HAI, Stanford Data Sciences and the Marlowe team, and the Google TPU Research Cloud team for computing support. 

% \newpage
\bibliography{babyview_references}

\begin{thebibliography}{}

\bibitem [\protect \citeauthoryear {%
Aslin%
}{%
Aslin%
}{%
{\protect \APACyear {2009}}%
}]{%
aslin2009infants}
\APACinsertmetastar {%
aslin2009infants}%
\begin{APACrefauthors}%
Aslin, R\BPBI N.%
\end{APACrefauthors}%
\unskip\
\newblock
\APACrefYearMonthDay{2009}{}{}.
\newblock
{\BBOQ}\APACrefatitle {How infants view natural scenes gathered from a head-mounted camera} {How infants view natural scenes gathered from a head-mounted camera}.{\BBCQ}
\newblock
\APACjournalVolNumPages{Optometry and Vision Science}{86}{6}{561--565}.
\PrintBackRefs{\CurrentBib}

\bibitem [\protect \citeauthoryear {%
Bergelson%
\ \BBA {} Aslin%
}{%
Bergelson%
\ \BBA {} Aslin%
}{%
{\protect \APACyear {2017}}%
}]{%
bergelson2017}
\APACinsertmetastar {%
bergelson2017}%
\begin{APACrefauthors}%
Bergelson, E.%
\BCBT {}\ \BBA {} Aslin, R\BPBI N.%
\end{APACrefauthors}%
\unskip\
\newblock
\APACrefYearMonthDay{2017}{}{}.
\newblock
{\BBOQ}\APACrefatitle {Nature and origins of the lexicon in 6-mo-olds} {Nature and origins of the lexicon in 6-mo-olds}.{\BBCQ}
\newblock
\APACjournalVolNumPages{Proceedings of the National Academy of Sciences}{114}{49}{12916--12921}.
\PrintBackRefs{\CurrentBib}

\bibitem [\protect \citeauthoryear {%
Brown%
\ \protect \BOthers {.}}{%
Brown%
\ \protect \BOthers {.}}{%
{\protect \APACyear {2020}}%
}]{%
brown2020}
\APACinsertmetastar {%
brown2020}%
\begin{APACrefauthors}%
Brown, T.%
, Mann, B.%
, Ryder, N.%
, Subbiah, M.%
, Kaplan, J\BPBI D.%
, Dhariwal, P.%
\BDBL {}others%
\end{APACrefauthors}%
\unskip\
\newblock
\APACrefYearMonthDay{2020}{}{}.
\newblock
{\BBOQ}\APACrefatitle {Language models are few-shot learners} {Language models are few-shot learners}.{\BBCQ}
\newblock
\APACjournalVolNumPages{Advances in neural information processing systems}{33}{}{1877--1901}.
\PrintBackRefs{\CurrentBib}

\bibitem [\protect \citeauthoryear {%
Caesar%
, Uijlings%
\BCBL {}\ \BBA {} Ferrari%
}{%
Caesar%
\ \protect \BOthers {.}}{%
{\protect \APACyear {2018}}%
}]{%
caesar2018coco}
\APACinsertmetastar {%
caesar2018coco}%
\begin{APACrefauthors}%
Caesar, H.%
, Uijlings, J.%
\BCBL {}\ \BBA {} Ferrari, V.%
\end{APACrefauthors}%
\unskip\
\newblock
\APACrefYearMonthDay{2018}{}{}.
\newblock
{\BBOQ}\APACrefatitle {Coco-stuff: Thing and stuff classes in context} {Coco-stuff: Thing and stuff classes in context}.{\BBCQ}
\newblock
\BIn{} \APACrefbtitle {Proceedings of the IEEE conference on computer vision and pattern recognition} {Proceedings of the ieee conference on computer vision and pattern recognition}\ (\BPGS\ 1209--1218).
\PrintBackRefs{\CurrentBib}

\bibitem [\protect \citeauthoryear {%
Carey%
\ \BBA {} Bartlett%
}{%
Carey%
\ \BBA {} Bartlett%
}{%
{\protect \APACyear {1978}}%
}]{%
carey1978acquiring}
\APACinsertmetastar {%
carey1978acquiring}%
\begin{APACrefauthors}%
Carey, S.%
\BCBT {}\ \BBA {} Bartlett, E.%
\end{APACrefauthors}%
\unskip\
\newblock
\APACrefYearMonthDay{1978}{}{}.
\newblock
{\BBOQ}\APACrefatitle {Acquiring a single new word.} {Acquiring a single new word.}{\BBCQ}
\newblock
\APACjournalVolNumPages{Linguistics}{}{}{}.
\PrintBackRefs{\CurrentBib}

\bibitem [\protect \citeauthoryear {%
Caron%
\ \protect \BOthers {.}}{%
Caron%
\ \protect \BOthers {.}}{%
{\protect \APACyear {2021}}%
}]{%
dinov1}
\APACinsertmetastar {%
dinov1}%
\begin{APACrefauthors}%
Caron, M.%
, Touvron, H.%
, Misra, I.%
, J{\'e}gou, H.%
, Mairal, J.%
, Bojanowski, P.%
\BCBL {}\ \BBA {} Joulin, A.%
\end{APACrefauthors}%
\unskip\
\newblock
\APACrefYearMonthDay{2021}{}{}.
\newblock
{\BBOQ}\APACrefatitle {Emerging properties in self-supervised vision transformers} {Emerging properties in self-supervised vision transformers}.{\BBCQ}
\newblock
\BIn{} \APACrefbtitle {Proceedings of the IEEE/CVF International Conference on Computer Vision} {Proceedings of the ieee/cvf international conference on computer vision}\ (\BPGS\ 9650--9660).
\PrintBackRefs{\CurrentBib}

\bibitem [\protect \citeauthoryear {%
T.~Chen%
, Kornblith%
, Norouzi%
\BCBL {}\ \BBA {} Hinton%
}{%
T.~Chen%
\ \protect \BOthers {.}}{%
{\protect \APACyear {2020}}%
}]{%
simclr}
\APACinsertmetastar {%
simclr}%
\begin{APACrefauthors}%
Chen, T.%
, Kornblith, S.%
, Norouzi, M.%
\BCBL {}\ \BBA {} Hinton, G.%
\end{APACrefauthors}%
\unskip\
\newblock
\APACrefYearMonthDay{2020}{}{}.
\newblock
{\BBOQ}\APACrefatitle {A simple framework for contrastive learning of visual representations} {A simple framework for contrastive learning of visual representations}.{\BBCQ}
\newblock
\BIn{} \APACrefbtitle {ICML.} {Icml.}
\PrintBackRefs{\CurrentBib}

\bibitem [\protect \citeauthoryear {%
X.~Chen%
, Xie%
\BCBL {}\ \BBA {} He%
}{%
X.~Chen%
\ \protect \BOthers {.}}{%
{\protect \APACyear {2021}}%
}]{%
chen2021empirical}
\APACinsertmetastar {%
chen2021empirical}%
\begin{APACrefauthors}%
Chen, X.%
, Xie, S.%
\BCBL {}\ \BBA {} He, K.%
\end{APACrefauthors}%
\unskip\
\newblock
\APACrefYearMonthDay{2021}{}{}.
\newblock
{\BBOQ}\APACrefatitle {An empirical study of training self-supervised vision transformers} {An empirical study of training self-supervised vision transformers}.{\BBCQ}
\newblock
\BIn{} \APACrefbtitle {Proceedings of the IEEE/CVF international conference on computer vision} {Proceedings of the ieee/cvf international conference on computer vision}\ (\BPGS\ 9640--9649).
\PrintBackRefs{\CurrentBib}

\bibitem [\protect \citeauthoryear {%
Clerkin%
, Hart%
, Rehg%
, Yu%
\BCBL {}\ \BBA {} Smith%
}{%
Clerkin%
\ \protect \BOthers {.}}{%
{\protect \APACyear {2017}}%
}]{%
clerkin2017real}
\APACinsertmetastar {%
clerkin2017real}%
\begin{APACrefauthors}%
Clerkin, E\BPBI M.%
, Hart, E.%
, Rehg, J\BPBI M.%
, Yu, C.%
\BCBL {}\ \BBA {} Smith, L\BPBI B.%
\end{APACrefauthors}%
\unskip\
\newblock
\APACrefYearMonthDay{2017}{}{}.
\newblock
{\BBOQ}\APACrefatitle {Real-world visual statistics and infants' first-learned object names} {Real-world visual statistics and infants' first-learned object names}.{\BBCQ}
\newblock
\APACjournalVolNumPages{Philosophical Transactions of the Royal Society B: Biological Sciences}{372}{1711}{20160055}.
\PrintBackRefs{\CurrentBib}

\bibitem [\protect \citeauthoryear {%
Clerkin%
\ \BBA {} Smith%
}{%
Clerkin%
\ \BBA {} Smith%
}{%
{\protect \APACyear {2022}}%
}]{%
clerkin2022real}
\APACinsertmetastar {%
clerkin2022real}%
\begin{APACrefauthors}%
Clerkin, E\BPBI M.%
\BCBT {}\ \BBA {} Smith, L\BPBI B.%
\end{APACrefauthors}%
\unskip\
\newblock
\APACrefYearMonthDay{2022}{}{}.
\newblock
{\BBOQ}\APACrefatitle {Real-world statistics at two timescales and a mechanism for infant learning of object names} {Real-world statistics at two timescales and a mechanism for infant learning of object names}.{\BBCQ}
\newblock
\APACjournalVolNumPages{Proceedings of the National Academy of Sciences}{119}{18}{e2123239119}.
\PrintBackRefs{\CurrentBib}

\bibitem [\protect \citeauthoryear {%
Contributors%
}{%
Contributors%
}{%
{\protect \APACyear {2020}}%
{\protect \APACexlab {{\protect \BCnt {1}}}}}]{%
mmseg2020}
\APACinsertmetastar {%
mmseg2020}%
\begin{APACrefauthors}%
Contributors, M.%
\end{APACrefauthors}%
\unskip\
\newblock
\APACrefYearMonthDay{2020{\protect \BCnt {1}}}{}{}.
\newblock
\APACrefbtitle {{MMSegmentation}: OpenMMLab Semantic Segmentation Toolbox and Benchmark.} {{MMSegmentation}: Openmmlab semantic segmentation toolbox and benchmark.}
\newblock
\APAChowpublished {\url{https://github.com/open-mmlab/mmsegmentation}}.
\PrintBackRefs{\CurrentBib}

\bibitem [\protect \citeauthoryear {%
Contributors%
}{%
Contributors%
}{%
{\protect \APACyear {2020}}%
{\protect \APACexlab {{\protect \BCnt {2}}}}}]{%
mmpose2020}
\APACinsertmetastar {%
mmpose2020}%
\begin{APACrefauthors}%
Contributors, M.%
\end{APACrefauthors}%
\unskip\
\newblock
\APACrefYearMonthDay{2020{\protect \BCnt {2}}}{}{}.
\newblock
\APACrefbtitle {OpenMMLab Pose Estimation Toolbox and Benchmark.} {Openmmlab pose estimation toolbox and benchmark.}
\newblock
\APAChowpublished {\url{https://github.com/open-mmlab/mmpose}}.
\PrintBackRefs{\CurrentBib}

\bibitem [\protect \citeauthoryear {%
Conwell%
, Prince%
, Kay%
, Alvarez%
\BCBL {}\ \BBA {} Konkle%
}{%
Conwell%
\ \protect \BOthers {.}}{%
{\protect \APACyear {2024}}%
}]{%
conwell2024large}
\APACinsertmetastar {%
conwell2024large}%
\begin{APACrefauthors}%
Conwell, C.%
, Prince, J\BPBI S.%
, Kay, K\BPBI N.%
, Alvarez, G\BPBI A.%
\BCBL {}\ \BBA {} Konkle, T.%
\end{APACrefauthors}%
\unskip\
\newblock
\APACrefYearMonthDay{2024}{}{}.
\newblock
{\BBOQ}\APACrefatitle {A large-scale examination of inductive biases shaping high-level visual representation in brains and machines} {A large-scale examination of inductive biases shaping high-level visual representation in brains and machines}.{\BBCQ}
\newblock
\APACjournalVolNumPages{Nature communications}{15}{1}{9383}.
\PrintBackRefs{\CurrentBib}

\bibitem [\protect \citeauthoryear {%
Damen%
\ \protect \BOthers {.}}{%
Damen%
\ \protect \BOthers {.}}{%
{\protect \APACyear {2018}}%
}]{%
damen2018scaling}
\APACinsertmetastar {%
damen2018scaling}%
\begin{APACrefauthors}%
Damen, D.%
, Doughty, H.%
, Farinella, G\BPBI M.%
, Fidler, S.%
, Furnari, A.%
, Kazakos, E.%
\BDBL {}others%
\end{APACrefauthors}%
\unskip\
\newblock
\APACrefYearMonthDay{2018}{}{}.
\newblock
{\BBOQ}\APACrefatitle {Scaling egocentric vision: The epic-kitchens dataset} {Scaling egocentric vision: The epic-kitchens dataset}.{\BBCQ}
\newblock
\BIn{} \APACrefbtitle {Proceedings of the European conference on computer vision (ECCV)} {Proceedings of the european conference on computer vision (eccv)}\ (\BPGS\ 720--736).
\PrintBackRefs{\CurrentBib}

\bibitem [\protect \citeauthoryear {%
Damen%
\ \protect \BOthers {.}}{%
Damen%
\ \protect \BOthers {.}}{%
{\protect \APACyear {2022}}%
}]{%
damen2022}
\APACinsertmetastar {%
damen2022}%
\begin{APACrefauthors}%
Damen, D.%
, Doughty, H.%
, Farinella, G\BPBI M.%
, Furnari, A.%
, Kazakos, E.%
, Ma, J.%
\BDBL {}others%
\end{APACrefauthors}%
\unskip\
\newblock
\APACrefYearMonthDay{2022}{}{}.
\newblock
{\BBOQ}\APACrefatitle {Rescaling egocentric vision: Collection, pipeline and challenges for epic-kitchens-100} {Rescaling egocentric vision: Collection, pipeline and challenges for epic-kitchens-100}.{\BBCQ}
\newblock
\APACjournalVolNumPages{International Journal of Computer Vision}{}{}{1--23}.
\PrintBackRefs{\CurrentBib}

\bibitem [\protect \citeauthoryear {%
de Barbaro%
\ \BBA {} Fausey%
}{%
de Barbaro%
\ \BBA {} Fausey%
}{%
{\protect \APACyear {2022}}%
}]{%
de2022ten}
\APACinsertmetastar {%
de2022ten}%
\begin{APACrefauthors}%
de Barbaro, K.%
\BCBT {}\ \BBA {} Fausey, C\BPBI M.%
\end{APACrefauthors}%
\unskip\
\newblock
\APACrefYearMonthDay{2022}{}{}.
\newblock
{\BBOQ}\APACrefatitle {Ten lessons about infants’ everyday experiences} {Ten lessons about infants’ everyday experiences}.{\BBCQ}
\newblock
\APACjournalVolNumPages{Current Directions in Psychological Science}{31}{1}{28--33}.
\PrintBackRefs{\CurrentBib}

\bibitem [\protect \citeauthoryear {%
Fausey%
, Jayaraman%
\BCBL {}\ \BBA {} Smith%
}{%
Fausey%
\ \protect \BOthers {.}}{%
{\protect \APACyear {2016}}%
}]{%
fausey2016faces}
\APACinsertmetastar {%
fausey2016faces}%
\begin{APACrefauthors}%
Fausey, C\BPBI M.%
, Jayaraman, S.%
\BCBL {}\ \BBA {} Smith, L\BPBI B.%
\end{APACrefauthors}%
\unskip\
\newblock
\APACrefYearMonthDay{2016}{}{}.
\newblock
{\BBOQ}\APACrefatitle {From faces to hands: Changing visual input in the first two years} {From faces to hands: Changing visual input in the first two years}.{\BBCQ}
\newblock
\APACjournalVolNumPages{Cognition}{152}{}{101--107}.
\PrintBackRefs{\CurrentBib}

\bibitem [\protect \citeauthoryear {%
Feng%
, Goodman%
\BCBL {}\ \BBA {} Frank%
}{%
Feng%
\ \protect \BOthers {.}}{%
{\protect \APACyear {2024}}%
}]{%
feng2024childdirectedspeecheffectivetraining}
\APACinsertmetastar {%
feng2024childdirectedspeecheffectivetraining}%
\begin{APACrefauthors}%
Feng, S\BPBI Y.%
, Goodman, N\BPBI D.%
\BCBL {}\ \BBA {} Frank, M\BPBI C.%
\end{APACrefauthors}%
\unskip\
\newblock
\APACrefYearMonthDay{2024}{}{}.
\newblock
{\BBOQ}\APACrefatitle {Is Child-Directed Speech Effective Training Data for Language Models?} {Is child-directed speech effective training data for language models?}{\BBCQ}
\newblock
\BIn{} \APACrefbtitle {Proceedings of the 2024 Conference on Empirical Methods in Natural Language Processing.} {Proceedings of the 2024 conference on empirical methods in natural language processing.}
\newblock
\APACaddressPublisher{}{Association for Computational Linguistics}.
\newblock
\begin{APACrefURL} \url{https://arxiv.org/abs/2408.03617} \end{APACrefURL}
\PrintBackRefs{\CurrentBib}

\bibitem [\protect \citeauthoryear {%
Franchak%
, Kretch%
, Soska%
\BCBL {}\ \BBA {} Adolph%
}{%
Franchak%
\ \protect \BOthers {.}}{%
{\protect \APACyear {2011}}%
}]{%
franchak2011}
\APACinsertmetastar {%
franchak2011}%
\begin{APACrefauthors}%
Franchak, J\BPBI M.%
, Kretch, K\BPBI S.%
, Soska, K\BPBI C.%
\BCBL {}\ \BBA {} Adolph, K\BPBI E.%
\end{APACrefauthors}%
\unskip\
\newblock
\APACrefYearMonthDay{2011}{}{}.
\newblock
{\BBOQ}\APACrefatitle {Head-mounted eye tracking: A new method to describe infant looking} {Head-mounted eye tracking: A new method to describe infant looking}.{\BBCQ}
\newblock
\APACjournalVolNumPages{Child development}{82}{6}{1738--1750}.
\PrintBackRefs{\CurrentBib}

\bibitem [\protect \citeauthoryear {%
Frank%
}{%
Frank%
}{%
{\protect \APACyear {2023}}%
}]{%
frank2023}
\APACinsertmetastar {%
frank2023}%
\begin{APACrefauthors}%
Frank, M\BPBI C.%
\end{APACrefauthors}%
\unskip\
\newblock
\APACrefYearMonthDay{2023}{}{}.
\newblock
{\BBOQ}\APACrefatitle {Bridging the data gap between children and large language models} {Bridging the data gap between children and large language models}.{\BBCQ}
\newblock
\APACjournalVolNumPages{Trends in Cognitive Sciences}{}{}{}.
\PrintBackRefs{\CurrentBib}

\bibitem [\protect \citeauthoryear {%
Frank%
, Braginsky%
, Yurovsky%
\BCBL {}\ \BBA {} Marchman%
}{%
Frank%
\ \protect \BOthers {.}}{%
{\protect \APACyear {2021}}%
}]{%
frank2021variability}
\APACinsertmetastar {%
frank2021variability}%
\begin{APACrefauthors}%
Frank, M\BPBI C.%
, Braginsky, M.%
, Yurovsky, D.%
\BCBL {}\ \BBA {} Marchman, V\BPBI A.%
\end{APACrefauthors}%
\unskip\
\newblock
\APACrefYear{2021}.
\newblock
\APACrefbtitle {Variability and consistency in early language learning: The Wordbank project} {Variability and consistency in early language learning: The wordbank project}.
\newblock
\APACaddressPublisher{}{MIT Press}.
\PrintBackRefs{\CurrentBib}

\bibitem [\protect \citeauthoryear {%
Gandhi%
, von Platen%
\BCBL {}\ \BBA {} Rush%
}{%
Gandhi%
\ \protect \BOthers {.}}{%
{\protect \APACyear {2023}}%
}]{%
gandhi2023distil}
\APACinsertmetastar {%
gandhi2023distil}%
\begin{APACrefauthors}%
Gandhi, S.%
, von Platen, P.%
\BCBL {}\ \BBA {} Rush, A\BPBI M.%
\end{APACrefauthors}%
\unskip\
\newblock
\APACrefYearMonthDay{2023}{}{}.
\newblock
{\BBOQ}\APACrefatitle {Distil-Whisper: Robust Knowledge Distillation via Large-Scale Pseudo Labelling} {Distil-whisper: Robust knowledge distillation via large-scale pseudo labelling}.{\BBCQ}
\newblock
\APACjournalVolNumPages{arXiv preprint arXiv:2311.00430}{}{}{}.
\PrintBackRefs{\CurrentBib}

\bibitem [\protect \citeauthoryear {%
Gilmore%
, Adolph%
\BCBL {}\ \BBA {} Millman%
}{%
Gilmore%
\ \protect \BOthers {.}}{%
{\protect \APACyear {2016}}%
}]{%
gilmore2016curating}
\APACinsertmetastar {%
gilmore2016curating}%
\begin{APACrefauthors}%
Gilmore, R\BPBI O.%
, Adolph, K\BPBI E.%
\BCBL {}\ \BBA {} Millman, D\BPBI S.%
\end{APACrefauthors}%
\unskip\
\newblock
\APACrefYearMonthDay{2016}{}{}.
\newblock
{\BBOQ}\APACrefatitle {Curating identifiable data for sharing: The databrary project} {Curating identifiable data for sharing: The databrary project}.{\BBCQ}
\newblock
\BIn{} \APACrefbtitle {2016 New York Scientific Data Summit (NYSDS)} {2016 new york scientific data summit (nysds)}\ (\BPGS\ 1--6).
\PrintBackRefs{\CurrentBib}

\bibitem [\protect \citeauthoryear {%
Grauman%
\ \protect \BOthers {.}}{%
Grauman%
\ \protect \BOthers {.}}{%
{\protect \APACyear {2022}}%
}]{%
grauman2022}
\APACinsertmetastar {%
grauman2022}%
\begin{APACrefauthors}%
Grauman, K.%
, Westbury, A.%
, Byrne, E.%
, Chavis, Z.%
, Furnari, A.%
, Girdhar, R.%
\BDBL {}others%
\end{APACrefauthors}%
\unskip\
\newblock
\APACrefYearMonthDay{2022}{}{}.
\newblock
{\BBOQ}\APACrefatitle {Ego4d: Around the world in 3,000 hours of egocentric video} {Ego4d: Around the world in 3,000 hours of egocentric video}.{\BBCQ}
\newblock
\BIn{} \APACrefbtitle {Proceedings of the IEEE/CVF Conference on Computer Vision and Pattern Recognition} {Proceedings of the ieee/cvf conference on computer vision and pattern recognition}\ (\BPGS\ 18995--19012).
\PrintBackRefs{\CurrentBib}

\bibitem [\protect \citeauthoryear {%
He%
\ \protect \BOthers {.}}{%
He%
\ \protect \BOthers {.}}{%
{\protect \APACyear {2021}}%
}]{%
mae}
\APACinsertmetastar {%
mae}%
\begin{APACrefauthors}%
He, K.%
, Chen, X.%
, Xie, S.%
, Li, Y.%
, Doll{\'a}r, P.%
\BCBL {}\ \BBA {} Girshick, R.%
\end{APACrefauthors}%
\unskip\
\newblock
\APACrefYearMonthDay{2021}{}{}.
\newblock
{\BBOQ}\APACrefatitle {Masked autoencoders are scalable vision learners} {Masked autoencoders are scalable vision learners}.{\BBCQ}
\newblock
\APACjournalVolNumPages{arXiv preprint arXiv:2111.06377}{}{}{}.
\PrintBackRefs{\CurrentBib}

\bibitem [\protect \citeauthoryear {%
He%
, Fan%
, Wu%
, Xie%
\BCBL {}\ \BBA {} Girshick%
}{%
He%
\ \protect \BOthers {.}}{%
{\protect \APACyear {2020}}%
}]{%
moco}
\APACinsertmetastar {%
moco}%
\begin{APACrefauthors}%
He, K.%
, Fan, H.%
, Wu, Y.%
, Xie, S.%
\BCBL {}\ \BBA {} Girshick, R.%
\end{APACrefauthors}%
\unskip\
\newblock
\APACrefYearMonthDay{2020}{}{}.
\newblock
{\BBOQ}\APACrefatitle {Momentum contrast for unsupervised visual representation learning} {Momentum contrast for unsupervised visual representation learning}.{\BBCQ}
\newblock
\BIn{} \APACrefbtitle {Proceedings of the IEEE/CVF Conference on Computer Vision and Pattern Recognition} {Proceedings of the ieee/cvf conference on computer vision and pattern recognition}\ (\BPGS\ 9729--9738).
\PrintBackRefs{\CurrentBib}

\bibitem [\protect \citeauthoryear {%
Huber%
, Geirhos%
\BCBL {}\ \BBA {} Wichmann%
}{%
Huber%
\ \protect \BOthers {.}}{%
{\protect \APACyear {2023}}%
}]{%
huber2023developmental}
\APACinsertmetastar {%
huber2023developmental}%
\begin{APACrefauthors}%
Huber, L\BPBI S.%
, Geirhos, R.%
\BCBL {}\ \BBA {} Wichmann, F\BPBI A.%
\end{APACrefauthors}%
\unskip\
\newblock
\APACrefYearMonthDay{2023}{}{}.
\newblock
{\BBOQ}\APACrefatitle {The developmental trajectory of object recognition robustness: children are like small adults but unlike big deep neural networks} {The developmental trajectory of object recognition robustness: children are like small adults but unlike big deep neural networks}.{\BBCQ}
\newblock
\APACjournalVolNumPages{Journal of vision}{23}{7}{4--4}.
\PrintBackRefs{\CurrentBib}

\bibitem [\protect \citeauthoryear {%
Huebner%
, Sulem%
, Cynthia%
\BCBL {}\ \BBA {} Roth%
}{%
Huebner%
\ \protect \BOthers {.}}{%
{\protect \APACyear {2021}}%
}]{%
huebner-etal-2021-babyberta}
\APACinsertmetastar {%
huebner-etal-2021-babyberta}%
\begin{APACrefauthors}%
Huebner, P\BPBI A.%
, Sulem, E.%
, Cynthia, F.%
\BCBL {}\ \BBA {} Roth, D.%
\end{APACrefauthors}%
\unskip\
\newblock
\APACrefYearMonthDay{2021}{{\APACmonth{11}}}{}.
\newblock
{\BBOQ}\APACrefatitle {{B}aby{BERT}a: Learning More Grammar With Small-Scale Child-Directed Language} {{B}aby{BERT}a: Learning more grammar with small-scale child-directed language}.{\BBCQ}
\newblock
\BIn{} A.~Bisazza\ \BBA {} O.~Abend\ (\BEDS), \APACrefbtitle {Proceedings of the 25th Conference on Computational Natural Language Learning} {Proceedings of the 25th conference on computational natural language learning}\ (\BPGS\ 624--646).
\newblock
\APACaddressPublisher{Online}{Association for Computational Linguistics}.
\newblock
\begin{APACrefURL} \url{https://aclanthology.org/2021.conll-1.49} \end{APACrefURL}
\newblock
\begin{APACrefDOI} \doi{10.18653/v1/2021.conll-1.49} \end{APACrefDOI}
\PrintBackRefs{\CurrentBib}

\bibitem [\protect \citeauthoryear {%
Jackson-Maldonado%
\ \protect \BOthers {.}}{%
Jackson-Maldonado%
\ \protect \BOthers {.}}{%
{\protect \APACyear {2003}}%
}]{%
jacksonmaldonado2003}
\APACinsertmetastar {%
jacksonmaldonado2003}%
\begin{APACrefauthors}%
Jackson-Maldonado, D.%
, Thal, D\BPBI J.%
, Fenson, L.%
, Marchman, V\BPBI A.%
, Newton, T.%
\BCBL {}\ \BBA {} Barbara, C.%
\end{APACrefauthors}%
\unskip\
\newblock
\APACrefYear{2003}.
\newblock
\APACrefbtitle {MacArthur-Bates Inventarios del Desarollo de Habilitades Communicativas: User's Guide and Technical Manual} {Macarthur-bates inventarios del desarollo de habilitades communicativas: User's guide and technical manual}.
\newblock
\APACaddressPublisher{}{Brookes Publishing Company}.
\PrintBackRefs{\CurrentBib}

\bibitem [\protect \citeauthoryear {%
Jiang%
\ \protect \BOthers {.}}{%
Jiang%
\ \protect \BOthers {.}}{%
{\protect \APACyear {2023}}%
}]{%
jiang2023rtmpose}
\APACinsertmetastar {%
jiang2023rtmpose}%
\begin{APACrefauthors}%
Jiang, T.%
, Lu, P.%
, Zhang, L.%
, Ma, N.%
, Han, R.%
, Lyu, C.%
\BDBL {}Chen, K.%
\end{APACrefauthors}%
\unskip\
\newblock
\APACrefYearMonthDay{2023}{}{}.
\newblock
{\BBOQ}\APACrefatitle {Rtmpose: Real-time multi-person pose estimation based on mmpose} {Rtmpose: Real-time multi-person pose estimation based on mmpose}.{\BBCQ}
\newblock
\APACjournalVolNumPages{arXiv preprint arXiv:2303.07399}{}{}{}.
\PrintBackRefs{\CurrentBib}

\bibitem [\protect \citeauthoryear {%
B.~Joshi%
, Xanthidis%
, Rahman%
\BCBL {}\ \BBA {} Rekleitis%
}{%
B.~Joshi%
\ \protect \BOthers {.}}{%
{\protect \APACyear {2022}}%
}]{%
joshi_gopro_icra_2022}
\APACinsertmetastar {%
joshi_gopro_icra_2022}%
\begin{APACrefauthors}%
Joshi, B.%
, Xanthidis, M.%
, Rahman, S.%
\BCBL {}\ \BBA {} Rekleitis, I.%
\end{APACrefauthors}%
\unskip\
\newblock
\APACrefYearMonthDay{2022}{}{}.
\newblock
{\BBOQ}\APACrefatitle {High Definition, Inexpensive, Underwater Mapping} {High definition, inexpensive, underwater mapping}.{\BBCQ}
\newblock
\BIn{} \APACrefbtitle {IEEE International Conference on Robotics and Automation (ICRA)} {Ieee international conference on robotics and automation (icra)}\ (\BPG~1113-1121).
\newblock
\begin{APACrefDOI} \doi{10.1109/ICRA46639.2022.9811695} \end{APACrefDOI}
\PrintBackRefs{\CurrentBib}

\bibitem [\protect \citeauthoryear {%
N.~Joshi%
, Kang%
, Zitnick%
\BCBL {}\ \BBA {} Szeliski%
}{%
N.~Joshi%
\ \protect \BOthers {.}}{%
{\protect \APACyear {2010}}%
}]{%
joshi2010image}
\APACinsertmetastar {%
joshi2010image}%
\begin{APACrefauthors}%
Joshi, N.%
, Kang, S\BPBI B.%
, Zitnick, C\BPBI L.%
\BCBL {}\ \BBA {} Szeliski, R.%
\end{APACrefauthors}%
\unskip\
\newblock
\APACrefYearMonthDay{2010}{}{}.
\newblock
{\BBOQ}\APACrefatitle {Image deblurring using inertial measurement sensors} {Image deblurring using inertial measurement sensors}.{\BBCQ}
\newblock
\APACjournalVolNumPages{ACM Transactions on Graphics (TOG)}{29}{4}{1--9}.
\PrintBackRefs{\CurrentBib}

\bibitem [\protect \citeauthoryear {%
Karpenko%
, Jacobs%
, Baek%
\BCBL {}\ \BBA {} Levoy%
}{%
Karpenko%
\ \protect \BOthers {.}}{%
{\protect \APACyear {2011}}%
}]{%
karpenko2011digital}
\APACinsertmetastar {%
karpenko2011digital}%
\begin{APACrefauthors}%
Karpenko, A.%
, Jacobs, D.%
, Baek, J.%
\BCBL {}\ \BBA {} Levoy, M.%
\end{APACrefauthors}%
\unskip\
\newblock
\APACrefYearMonthDay{2011}{}{}.
\newblock
{\BBOQ}\APACrefatitle {Digital video stabilization and rolling shutter correction using gyroscopes} {Digital video stabilization and rolling shutter correction using gyroscopes}.{\BBCQ}
\newblock
\APACjournalVolNumPages{CSTR}{1}{2}{13}.
\PrintBackRefs{\CurrentBib}

\bibitem [\protect \citeauthoryear {%
Kretch%
, Franchak%
\BCBL {}\ \BBA {} Adolph%
}{%
Kretch%
\ \protect \BOthers {.}}{%
{\protect \APACyear {2014}}%
}]{%
kretch2014}
\APACinsertmetastar {%
kretch2014}%
\begin{APACrefauthors}%
Kretch, K\BPBI S.%
, Franchak, J\BPBI M.%
\BCBL {}\ \BBA {} Adolph, K\BPBI E.%
\end{APACrefauthors}%
\unskip\
\newblock
\APACrefYearMonthDay{2014}{}{}.
\newblock
{\BBOQ}\APACrefatitle {Crawling and walking infants see the world differently} {Crawling and walking infants see the world differently}.{\BBCQ}
\newblock
\APACjournalVolNumPages{Child development}{85}{4}{1503--1518}.
\PrintBackRefs{\CurrentBib}

\bibitem [\protect \citeauthoryear {%
Krizhevsky%
, Sutskever%
\BCBL {}\ \BBA {} Hinton%
}{%
Krizhevsky%
\ \protect \BOthers {.}}{%
{\protect \APACyear {2012}}%
}]{%
krizhevsky2012}
\APACinsertmetastar {%
krizhevsky2012}%
\begin{APACrefauthors}%
Krizhevsky, A.%
, Sutskever, I.%
\BCBL {}\ \BBA {} Hinton, G\BPBI E.%
\end{APACrefauthors}%
\unskip\
\newblock
\APACrefYearMonthDay{2012}{}{}.
\newblock
{\BBOQ}\APACrefatitle {Imagenet classification with deep convolutional neural networks} {Imagenet classification with deep convolutional neural networks}.{\BBCQ}
\newblock
\APACjournalVolNumPages{Advances in neural information processing systems}{25}{}{}.
\PrintBackRefs{\CurrentBib}

\bibitem [\protect \citeauthoryear {%
Lavechin%
, Bousbib%
, Bredin%
, Dupoux%
\BCBL {}\ \BBA {} Cristia%
}{%
Lavechin%
\ \protect \BOthers {.}}{%
{\protect \APACyear {2020}}%
}]{%
lavechin2020open}
\APACinsertmetastar {%
lavechin2020open}%
\begin{APACrefauthors}%
Lavechin, M.%
, Bousbib, R.%
, Bredin, H.%
, Dupoux, E.%
\BCBL {}\ \BBA {} Cristia, A.%
\end{APACrefauthors}%
\unskip\
\newblock
\APACrefYearMonthDay{2020}{}{}.
\newblock
{\BBOQ}\APACrefatitle {An open-source voice type classifier for child-centered daylong recordings} {An open-source voice type classifier for child-centered daylong recordings}.{\BBCQ}
\newblock
\APACjournalVolNumPages{arXiv preprint arXiv:2005.12656}{}{}{}.
\PrintBackRefs{\CurrentBib}

\bibitem [\protect \citeauthoryear {%
Y.~Li%
\ \protect \BOthers {.}}{%
Y.~Li%
\ \protect \BOthers {.}}{%
{\protect \APACyear {2022}}%
}]{%
li2022simcc}
\APACinsertmetastar {%
li2022simcc}%
\begin{APACrefauthors}%
Li, Y.%
, Yang, S.%
, Liu, P.%
, Zhang, S.%
, Wang, Y.%
, Wang, Z.%
\BDBL {}Xia, S\BHBI T.%
\end{APACrefauthors}%
\unskip\
\newblock
\APACrefYearMonthDay{2022}{}{}.
\newblock
{\BBOQ}\APACrefatitle {Simcc: A simple coordinate classification perspective for human pose estimation} {Simcc: A simple coordinate classification perspective for human pose estimation}.{\BBCQ}
\newblock
\BIn{} \APACrefbtitle {European Conference on Computer Vision} {European conference on computer vision}\ (\BPGS\ 89--106).
\PrintBackRefs{\CurrentBib}

\bibitem [\protect \citeauthoryear {%
Z.~Li%
}{%
Z.~Li%
}{%
{\protect \APACyear {2022}}%
}]{%
lidepthtoolbox2022}
\APACinsertmetastar {%
lidepthtoolbox2022}%
\begin{APACrefauthors}%
Li, Z.%
\end{APACrefauthors}%
\unskip\
\newblock
\APACrefYearMonthDay{2022}{}{}.
\newblock
\APACrefbtitle {Monocular Depth Estimation Toolbox.} {Monocular depth estimation toolbox.}
\newblock
\APAChowpublished {\url{https://github.com/zhyever/Monocular-Depth-Estimation-Toolbox}}.
\PrintBackRefs{\CurrentBib}

\bibitem [\protect \citeauthoryear {%
Lin%
\ \protect \BOthers {.}}{%
Lin%
\ \protect \BOthers {.}}{%
{\protect \APACyear {2014}}%
}]{%
lin2014microsoft}
\APACinsertmetastar {%
lin2014microsoft}%
\begin{APACrefauthors}%
Lin, T\BHBI Y.%
, Maire, M.%
, Belongie, S.%
, Hays, J.%
, Perona, P.%
, Ramanan, D.%
\BDBL {}Zitnick, C\BPBI L.%
\end{APACrefauthors}%
\unskip\
\newblock
\APACrefYearMonthDay{2014}{}{}.
\newblock
{\BBOQ}\APACrefatitle {Microsoft coco: Common objects in context} {Microsoft coco: Common objects in context}.{\BBCQ}
\newblock
\BIn{} \APACrefbtitle {Computer Vision--ECCV 2014: 13th European Conference, Zurich, Switzerland, September 6-12, 2014, Proceedings, Part V 13} {Computer vision--eccv 2014: 13th european conference, zurich, switzerland, september 6-12, 2014, proceedings, part v 13}\ (\BPGS\ 740--755).
\PrintBackRefs{\CurrentBib}

\bibitem [\protect \citeauthoryear {%
Liu%
\ \protect \BOthers {.}}{%
Liu%
\ \protect \BOthers {.}}{%
{\protect \APACyear {2019}}%
}]{%
liu2019robertarobustlyoptimizedbert}
\APACinsertmetastar {%
liu2019robertarobustlyoptimizedbert}%
\begin{APACrefauthors}%
Liu, Y.%
, Ott, M.%
, Goyal, N.%
, Du, J.%
, Joshi, M.%
, Chen, D.%
\BDBL {}Stoyanov, V.%
\end{APACrefauthors}%
\unskip\
\newblock
\APACrefYearMonthDay{2019}{}{}.
\newblock
\APACrefbtitle {RoBERTa: A Robustly Optimized BERT Pretraining Approach.} {Roberta: A robustly optimized bert pretraining approach.}
\newblock
\begin{APACrefURL} \url{https://arxiv.org/abs/1907.11692} \end{APACrefURL}
\PrintBackRefs{\CurrentBib}

\bibitem [\protect \citeauthoryear {%
Long%
, Fan%
, Huey%
, Chai%
\BCBL {}\ \BBA {} Frank%
}{%
Long%
\ \protect \BOthers {.}}{%
{\protect \APACyear {2024}}%
}]{%
long2024parallel}
\APACinsertmetastar {%
long2024parallel}%
\begin{APACrefauthors}%
Long, B.%
, Fan, J\BPBI E.%
, Huey, H.%
, Chai, Z.%
\BCBL {}\ \BBA {} Frank, M\BPBI C.%
\end{APACrefauthors}%
\unskip\
\newblock
\APACrefYearMonthDay{2024}{}{}.
\newblock
{\BBOQ}\APACrefatitle {Parallel developmental changes in children’s production and recognition of line drawings of visual concepts} {Parallel developmental changes in children’s production and recognition of line drawings of visual concepts}.{\BBCQ}
\newblock
\APACjournalVolNumPages{Nature Communications}{15}{1}{1191}.
\PrintBackRefs{\CurrentBib}

\bibitem [\protect \citeauthoryear {%
Long%
\ \protect \BOthers {.}}{%
Long%
\ \protect \BOthers {.}}{%
{\protect \APACyear {2023}}%
}]{%
long2023}
\APACinsertmetastar {%
long2023}%
\begin{APACrefauthors}%
Long, B.%
, Goodin, S.%
, Kachergis, G.%
, Marchman, V\BPBI A.%
, Radwan, S\BPBI F.%
, Sparks, R\BPBI Z.%
\BDBL {}others%
\end{APACrefauthors}%
\unskip\
\newblock
\APACrefYearMonthDay{2023}{}{}.
\newblock
{\BBOQ}\APACrefatitle {The BabyView camera: Designing a new head-mounted camera to capture children's early social and visual environments} {The babyview camera: Designing a new head-mounted camera to capture children's early social and visual environments}.{\BBCQ}
\newblock
\APACjournalVolNumPages{Behavior Research Methods}{}{}{1--12}.
\PrintBackRefs{\CurrentBib}

\bibitem [\protect \citeauthoryear {%
Long%
, Kachergis%
, Agrawal%
\BCBL {}\ \BBA {} Frank%
}{%
Long%
, Kachergis%
\BCBL {}\ \protect \BOthers {.}}{%
{\protect \APACyear {2022}}%
}]{%
long2022b}
\APACinsertmetastar {%
long2022b}%
\begin{APACrefauthors}%
Long, B.%
, Kachergis, G.%
, Agrawal, K.%
\BCBL {}\ \BBA {} Frank, M\BPBI C.%
\end{APACrefauthors}%
\unskip\
\newblock
\APACrefYearMonthDay{2022}{}{}.
\newblock
{\BBOQ}\APACrefatitle {A longitudinal analysis of the social information in infants' naturalistic visual experience using automated detections} {A longitudinal analysis of the social information in infants' naturalistic visual experience using automated detections}.{\BBCQ}
\newblock
\APACjournalVolNumPages{Developmental Psychology}{58}{}{2211-2229}.
\newblock
\begin{APACrefDOI} \doi{10.1037/dev0001414} \end{APACrefDOI}
\PrintBackRefs{\CurrentBib}

\bibitem [\protect \citeauthoryear {%
Long%
, Sanchez%
, Kraus%
, Agrawal%
\BCBL {}\ \BBA {} Frank%
}{%
Long%
, Sanchez%
\BCBL {}\ \protect \BOthers {.}}{%
{\protect \APACyear {2022}}%
}]{%
long2022a}
\APACinsertmetastar {%
long2022a}%
\begin{APACrefauthors}%
Long, B.%
, Sanchez, A.%
, Kraus, A\BPBI M.%
, Agrawal, K.%
\BCBL {}\ \BBA {} Frank, M\BPBI C.%
\end{APACrefauthors}%
\unskip\
\newblock
\APACrefYearMonthDay{2022}{}{}.
\newblock
{\BBOQ}\APACrefatitle {Automated detections reveal the social information in the changing infant view} {Automated detections reveal the social information in the changing infant view}.{\BBCQ}
\newblock
\APACjournalVolNumPages{Child Development}{93}{1}{101--116}.
\PrintBackRefs{\CurrentBib}

\bibitem [\protect \citeauthoryear {%
Lu%
\ \protect \BOthers {.}}{%
Lu%
\ \protect \BOthers {.}}{%
{\protect \APACyear {2023}}%
}]{%
lu2023rtmo}
\APACinsertmetastar {%
lu2023rtmo}%
\begin{APACrefauthors}%
Lu, P.%
, Jiang, T.%
, Li, Y.%
, Li, X.%
, Chen, K.%
\BCBL {}\ \BBA {} Yang, W.%
\end{APACrefauthors}%
\unskip\
\newblock
\APACrefYearMonthDay{2023}{}{}.
\newblock
{\BBOQ}\APACrefatitle {RTMO: Towards High-Performance One-Stage Real-Time Multi-Person Pose Estimation} {Rtmo: Towards high-performance one-stage real-time multi-person pose estimation}.{\BBCQ}
\newblock
\APACjournalVolNumPages{arXiv preprint arXiv:2312.07526}{}{}{}.
\PrintBackRefs{\CurrentBib}

\bibitem [\protect \citeauthoryear {%
MacWhinney%
}{%
MacWhinney%
}{%
{\protect \APACyear {2014}}%
}]{%
macwhinney2014childes}
\APACinsertmetastar {%
macwhinney2014childes}%
\begin{APACrefauthors}%
MacWhinney, B.%
\end{APACrefauthors}%
\unskip\
\newblock
\APACrefYear{2014}.
\newblock
\APACrefbtitle {The CHILDES project: Tools for analyzing talk, Volume I: Transcription format and programs} {The childes project: Tools for analyzing talk, volume i: Transcription format and programs}.
\newblock
\APACaddressPublisher{}{Psychology Press}.
\PrintBackRefs{\CurrentBib}

\bibitem [\protect \citeauthoryear {%
Maji%
, Nagori%
, Mathew%
\BCBL {}\ \BBA {} Poddar%
}{%
Maji%
\ \protect \BOthers {.}}{%
{\protect \APACyear {2022}}%
}]{%
maji2022yolo}
\APACinsertmetastar {%
maji2022yolo}%
\begin{APACrefauthors}%
Maji, D.%
, Nagori, S.%
, Mathew, M.%
\BCBL {}\ \BBA {} Poddar, D.%
\end{APACrefauthors}%
\unskip\
\newblock
\APACrefYearMonthDay{2022}{}{}.
\newblock
{\BBOQ}\APACrefatitle {Yolo-pose: Enhancing yolo for multi person pose estimation using object keypoint similarity loss} {Yolo-pose: Enhancing yolo for multi person pose estimation using object keypoint similarity loss}.{\BBCQ}
\newblock
\BIn{} \APACrefbtitle {Proceedings of the IEEE/CVF Conference on Computer Vision and Pattern Recognition} {Proceedings of the ieee/cvf conference on computer vision and pattern recognition}\ (\BPGS\ 2637--2646).
\PrintBackRefs{\CurrentBib}

\bibitem [\protect \citeauthoryear {%
Marchman%
, Dale%
\BCBL {}\ \BBA {} Fenson%
}{%
Marchman%
\ \protect \BOthers {.}}{%
{\protect \APACyear {2023}}%
}]{%
marchman2023}
\APACinsertmetastar {%
marchman2023}%
\begin{APACrefauthors}%
Marchman, V\BPBI A.%
, Dale, P\BPBI S.%
\BCBL {}\ \BBA {} Fenson, L.%
\end{APACrefauthors}%
\unskip\
\newblock
\APACrefYear{2023}.
\newblock
\APACrefbtitle {The MacArthur-Bates Communicative Development Inventories: User's Guide and Technical Manual, 3rd Edition} {The macarthur-bates communicative development inventories: User's guide and technical manual, 3rd edition}.
\newblock
\APACaddressPublisher{}{Brookes Publishing Company}.
\PrintBackRefs{\CurrentBib}

\bibitem [\protect \citeauthoryear {%
Oquab%
\ \protect \BOthers {.}}{%
Oquab%
\ \protect \BOthers {.}}{%
{\protect \APACyear {2023}}%
}]{%
oquab2023dinov2}
\APACinsertmetastar {%
oquab2023dinov2}%
\begin{APACrefauthors}%
Oquab, M.%
, Darcet, T.%
, Moutakanni, T.%
, Vo, H.%
, Szafraniec, M.%
, Khalidov, V.%
\BDBL {}others%
\end{APACrefauthors}%
\unskip\
\newblock
\APACrefYearMonthDay{2023}{}{}.
\newblock
{\BBOQ}\APACrefatitle {Dinov2: Learning robust visual features without supervision} {Dinov2: Learning robust visual features without supervision}.{\BBCQ}
\newblock
\APACjournalVolNumPages{arXiv preprint arXiv:2304.07193}{}{}{}.
\PrintBackRefs{\CurrentBib}

\bibitem [\protect \citeauthoryear {%
Orhan%
}{%
Orhan%
}{%
{\protect \APACyear {2021}}%
}]{%
orhan2021much}
\APACinsertmetastar {%
orhan2021much}%
\begin{APACrefauthors}%
Orhan, A\BPBI E.%
\end{APACrefauthors}%
\unskip\
\newblock
\APACrefYearMonthDay{2021}{}{}.
\newblock
{\BBOQ}\APACrefatitle {How much human-like visual experience do current self-supervised learning algorithms need in order to achieve human-level object recognition?} {How much human-like visual experience do current self-supervised learning algorithms need in order to achieve human-level object recognition?}{\BBCQ}
\newblock
\APACjournalVolNumPages{arXiv preprint arXiv:2109.11523}{}{}{}.
\PrintBackRefs{\CurrentBib}

\bibitem [\protect \citeauthoryear {%
Orhan%
}{%
Orhan%
}{%
{\protect \APACyear {2023}}%
}]{%
orhan2023scaling}
\APACinsertmetastar {%
orhan2023scaling}%
\begin{APACrefauthors}%
Orhan, A\BPBI E.%
\end{APACrefauthors}%
\unskip\
\newblock
\APACrefYearMonthDay{2023}{}{}.
\newblock
{\BBOQ}\APACrefatitle {Scaling may be all you need for achieving human-level object recognition capacity with human-like visual experience} {Scaling may be all you need for achieving human-level object recognition capacity with human-like visual experience}.{\BBCQ}
\newblock
\APACjournalVolNumPages{arXiv preprint arXiv:2308.03712}{}{}{}.
\PrintBackRefs{\CurrentBib}

\bibitem [\protect \citeauthoryear {%
Orhan%
, Gupta%
\BCBL {}\ \BBA {} Lake%
}{%
Orhan%
\ \protect \BOthers {.}}{%
{\protect \APACyear {2020}}%
}]{%
orhan2020self}
\APACinsertmetastar {%
orhan2020self}%
\begin{APACrefauthors}%
Orhan, A\BPBI E.%
, Gupta, V.%
\BCBL {}\ \BBA {} Lake, B\BPBI M.%
\end{APACrefauthors}%
\unskip\
\newblock
\APACrefYearMonthDay{2020}{}{}.
\newblock
{\BBOQ}\APACrefatitle {Self-supervised learning through the eyes of a child} {Self-supervised learning through the eyes of a child}.{\BBCQ}
\newblock
\APACjournalVolNumPages{Advances in Neural Information Processing Systems}{33}{}{}.
\PrintBackRefs{\CurrentBib}

\bibitem [\protect \citeauthoryear {%
Orhan%
\ \BBA {} Lake%
}{%
Orhan%
\ \BBA {} Lake%
}{%
{\protect \APACyear {2024}}%
}]{%
orhan2024learning}
\APACinsertmetastar {%
orhan2024learning}%
\begin{APACrefauthors}%
Orhan, A\BPBI E.%
\BCBT {}\ \BBA {} Lake, B\BPBI M.%
\end{APACrefauthors}%
\unskip\
\newblock
\APACrefYearMonthDay{2024}{}{}.
\newblock
{\BBOQ}\APACrefatitle {Learning high-level visual representations from a child’s perspective without strong inductive biases} {Learning high-level visual representations from a child’s perspective without strong inductive biases}.{\BBCQ}
\newblock
\APACjournalVolNumPages{Nature Machine Intelligence}{6}{3}{271--283}.
\PrintBackRefs{\CurrentBib}

\bibitem [\protect \citeauthoryear {%
Orhan%
, Wang%
, Wang%
, Ren%
\BCBL {}\ \BBA {} Lake%
}{%
Orhan%
\ \protect \BOthers {.}}{%
{\protect \APACyear {2024}}%
}]{%
orhan2024self}
\APACinsertmetastar {%
orhan2024self}%
\begin{APACrefauthors}%
Orhan, A\BPBI E.%
, Wang, W.%
, Wang, A\BPBI N.%
, Ren, M.%
\BCBL {}\ \BBA {} Lake, B\BPBI M.%
\end{APACrefauthors}%
\unskip\
\newblock
\APACrefYearMonthDay{2024}{}{}.
\newblock
{\BBOQ}\APACrefatitle {Self-supervised learning of video representations from a child's perspective} {Self-supervised learning of video representations from a child's perspective}.{\BBCQ}
\newblock
\APACjournalVolNumPages{arXiv preprint arXiv:2402.00300}{}{}{}.
\PrintBackRefs{\CurrentBib}

\bibitem [\protect \citeauthoryear {%
Qin%
, Wang%
\BCBL {}\ \BBA {} Lake%
}{%
Qin%
\ \protect \BOthers {.}}{%
{\protect \APACyear {2024}}%
}]{%
qin2024systematic}
\APACinsertmetastar {%
qin2024systematic}%
\begin{APACrefauthors}%
Qin, Y.%
, Wang, W.%
\BCBL {}\ \BBA {} Lake, B\BPBI M.%
\end{APACrefauthors}%
\unskip\
\newblock
\APACrefYearMonthDay{2024}{}{}.
\newblock
{\BBOQ}\APACrefatitle {A systematic investigation of learnability from single child linguistic input} {A systematic investigation of learnability from single child linguistic input}.{\BBCQ}
\newblock
\BIn{} \APACrefbtitle {Proceedings of the 46th Annual Conference of the Cognitive Science Society.} {Proceedings of the 46th annual conference of the cognitive science society.}
\PrintBackRefs{\CurrentBib}

\bibitem [\protect \citeauthoryear {%
Radford%
\ \protect \BOthers {.}}{%
Radford%
\ \protect \BOthers {.}}{%
{\protect \APACyear {2023}}%
}]{%
radford2023robust}
\APACinsertmetastar {%
radford2023robust}%
\begin{APACrefauthors}%
Radford, A.%
, Kim, J\BPBI W.%
, Xu, T.%
, Brockman, G.%
, McLeavey, C.%
\BCBL {}\ \BBA {} Sutskever, I.%
\end{APACrefauthors}%
\unskip\
\newblock
\APACrefYearMonthDay{2023}{}{}.
\newblock
{\BBOQ}\APACrefatitle {Robust speech recognition via large-scale weak supervision} {Robust speech recognition via large-scale weak supervision}.{\BBCQ}
\newblock
\BIn{} \APACrefbtitle {International Conference on Machine Learning} {International conference on machine learning}\ (\BPGS\ 28492--28518).
\PrintBackRefs{\CurrentBib}

\bibitem [\protect \citeauthoryear {%
Radford%
\ \protect \BOthers {.}}{%
Radford%
\ \protect \BOthers {.}}{%
{\protect \APACyear {2019}}%
}]{%
radford2019language}
\APACinsertmetastar {%
radford2019language}%
\begin{APACrefauthors}%
Radford, A.%
, Wu, J.%
, Child, R.%
, Luan, D.%
, Amodei, D.%
, Sutskever, I.%
\BCBL {}\ \BOthersPeriod {.}\end{APACrefauthors}%
\unskip\
\newblock
\APACrefYearMonthDay{2019}{}{}.
\newblock
{\BBOQ}\APACrefatitle {Language models are unsupervised multitask learners} {Language models are unsupervised multitask learners}.{\BBCQ}
\newblock
\APACjournalVolNumPages{OpenAI blog}{1}{8}{9}.
\PrintBackRefs{\CurrentBib}

\bibitem [\protect \citeauthoryear {%
Ranftl%
, Bochkovskiy%
\BCBL {}\ \BBA {} Koltun%
}{%
Ranftl%
\ \protect \BOthers {.}}{%
{\protect \APACyear {2021}}%
}]{%
ranftl2021vision}
\APACinsertmetastar {%
ranftl2021vision}%
\begin{APACrefauthors}%
Ranftl, R.%
, Bochkovskiy, A.%
\BCBL {}\ \BBA {} Koltun, V.%
\end{APACrefauthors}%
\unskip\
\newblock
\APACrefYearMonthDay{2021}{}{}.
\newblock
{\BBOQ}\APACrefatitle {Vision transformers for dense prediction} {Vision transformers for dense prediction}.{\BBCQ}
\newblock
\BIn{} \APACrefbtitle {Proceedings of the IEEE/CVF international conference on computer vision} {Proceedings of the ieee/cvf international conference on computer vision}\ (\BPGS\ 12179--12188).
\PrintBackRefs{\CurrentBib}

\bibitem [\protect \citeauthoryear {%
Russakovsky%
\ \protect \BOthers {.}}{%
Russakovsky%
\ \protect \BOthers {.}}{%
{\protect \APACyear {2015}}%
}]{%
russakovsky2015imagenet}
\APACinsertmetastar {%
russakovsky2015imagenet}%
\begin{APACrefauthors}%
Russakovsky, O.%
, Deng, J.%
, Su, H.%
, Krause, J.%
, Satheesh, S.%
, Ma, S.%
\BDBL {}others%
\end{APACrefauthors}%
\unskip\
\newblock
\APACrefYearMonthDay{2015}{}{}.
\newblock
{\BBOQ}\APACrefatitle {Imagenet large scale visual recognition challenge} {Imagenet large scale visual recognition challenge}.{\BBCQ}
\newblock
\APACjournalVolNumPages{International journal of computer vision}{115}{}{211--252}.
\PrintBackRefs{\CurrentBib}

\bibitem [\protect \citeauthoryear {%
Sheybani%
, Hansaria%
, Wood%
, Smith%
\BCBL {}\ \BBA {} Tiganj%
}{%
Sheybani%
\ \protect \BOthers {.}}{%
{\protect \APACyear {2024}}%
}]{%
sheybani2024curriculum}
\APACinsertmetastar {%
sheybani2024curriculum}%
\begin{APACrefauthors}%
Sheybani, S.%
, Hansaria, H.%
, Wood, J.%
, Smith, L.%
\BCBL {}\ \BBA {} Tiganj, Z.%
\end{APACrefauthors}%
\unskip\
\newblock
\APACrefYearMonthDay{2024}{}{}.
\newblock
{\BBOQ}\APACrefatitle {Curriculum Learning With Infant Egocentric Videos} {Curriculum learning with infant egocentric videos}.{\BBCQ}
\newblock
\APACjournalVolNumPages{Advances in Neural Information Processing Systems}{36}{}{}.
\PrintBackRefs{\CurrentBib}

\bibitem [\protect \citeauthoryear {%
Silberman%
, Hoiem%
, Kohli%
\BCBL {}\ \BBA {} Fergus%
}{%
Silberman%
\ \protect \BOthers {.}}{%
{\protect \APACyear {2012}}%
}]{%
silberman2012indoor}
\APACinsertmetastar {%
silberman2012indoor}%
\begin{APACrefauthors}%
Silberman, N.%
, Hoiem, D.%
, Kohli, P.%
\BCBL {}\ \BBA {} Fergus, R.%
\end{APACrefauthors}%
\unskip\
\newblock
\APACrefYearMonthDay{2012}{}{}.
\newblock
{\BBOQ}\APACrefatitle {Indoor segmentation and support inference from rgbd images} {Indoor segmentation and support inference from rgbd images}.{\BBCQ}
\newblock
\BIn{} \APACrefbtitle {Computer Vision--ECCV 2012: 12th European Conference on Computer Vision, Florence, Italy, October 7-13, 2012, Proceedings, Part V 12} {Computer vision--eccv 2012: 12th european conference on computer vision, florence, italy, october 7-13, 2012, proceedings, part v 12}\ (\BPGS\ 746--760).
\PrintBackRefs{\CurrentBib}

\bibitem [\protect \citeauthoryear {%
Smith%
\ \BBA {} Slone%
}{%
Smith%
\ \BBA {} Slone%
}{%
{\protect \APACyear {2017}}%
}]{%
smith2017developmental}
\APACinsertmetastar {%
smith2017developmental}%
\begin{APACrefauthors}%
Smith, L\BPBI B.%
\BCBT {}\ \BBA {} Slone, L\BPBI K.%
\end{APACrefauthors}%
\unskip\
\newblock
\APACrefYearMonthDay{2017}{}{}.
\newblock
{\BBOQ}\APACrefatitle {A developmental approach to machine learning?} {A developmental approach to machine learning?}{\BBCQ}
\newblock
\APACjournalVolNumPages{Frontiers in psychology}{8}{}{296143}.
\PrintBackRefs{\CurrentBib}

\bibitem [\protect \citeauthoryear {%
Smith%
, Yu%
, Yoshida%
\BCBL {}\ \BBA {} Fausey%
}{%
Smith%
\ \protect \BOthers {.}}{%
{\protect \APACyear {2015}}%
}]{%
smith2015}
\APACinsertmetastar {%
smith2015}%
\begin{APACrefauthors}%
Smith, L\BPBI B.%
, Yu, C.%
, Yoshida, H.%
\BCBL {}\ \BBA {} Fausey, C\BPBI M.%
\end{APACrefauthors}%
\unskip\
\newblock
\APACrefYearMonthDay{2015}{}{}.
\newblock
{\BBOQ}\APACrefatitle {Contributions of head-mounted cameras to studying the visual environments of infants and young children} {Contributions of head-mounted cameras to studying the visual environments of infants and young children}.{\BBCQ}
\newblock
\APACjournalVolNumPages{Journal of Cognition and Development}{16}{3}{407--419}.
\PrintBackRefs{\CurrentBib}

\bibitem [\protect \citeauthoryear {%
Sparks%
\ \protect \BOthers {.}}{%
Sparks%
\ \protect \BOthers {.}}{%
{\protect \APACyear {2024}}%
}]{%
sparks2024preschool}
\APACinsertmetastar {%
sparks2024preschool}%
\begin{APACrefauthors}%
Sparks, R\BPBI Z.%
, Long, B.%
, Keene, G\BPBI E.%
, Perez, M\BPBI J.%
, Tan, A\BPBI W.%
, Marchman, V\BPBI A.%
\BCBL {}\ \BBA {} Frank, M\BPBI C.%
\end{APACrefauthors}%
\unskip\
\newblock
\APACrefYearMonthDay{2024}{}{}.
\newblock
{\BBOQ}\APACrefatitle {Characterizing Contextual Variation in Children’s Preschool Language Environment Using Naturalistic Egocentric Videos} {Characterizing contextual variation in children’s preschool language environment using naturalistic egocentric videos}.{\BBCQ}
\newblock
\BIn{} \APACrefbtitle {Proceedings of the 46th Annual Conference of the Cognitive Science Society.} {Proceedings of the 46th annual conference of the cognitive science society.}
\PrintBackRefs{\CurrentBib}

\bibitem [\protect \citeauthoryear {%
Sullivan%
, Mei%
, Perfors%
, Wojcik%
\BCBL {}\ \BBA {} Frank%
}{%
Sullivan%
\ \protect \BOthers {.}}{%
{\protect \APACyear {2021}}%
}]{%
sullivan2021}
\APACinsertmetastar {%
sullivan2021}%
\begin{APACrefauthors}%
Sullivan, J.%
, Mei, M.%
, Perfors, A.%
, Wojcik, E.%
\BCBL {}\ \BBA {} Frank, M\BPBI C.%
\end{APACrefauthors}%
\unskip\
\newblock
\APACrefYearMonthDay{2021}{}{}.
\newblock
{\BBOQ}\APACrefatitle {SAYCam: A large, longitudinal audiovisual dataset recorded from the infant's perspective} {Saycam: A large, longitudinal audiovisual dataset recorded from the infant's perspective}.{\BBCQ}
\newblock
\APACjournalVolNumPages{Open mind}{5}{}{20--29}.
\PrintBackRefs{\CurrentBib}

\bibitem [\protect \citeauthoryear {%
Sun%
, Xiao%
, Liu%
\BCBL {}\ \BBA {} Wang%
}{%
Sun%
\ \protect \BOthers {.}}{%
{\protect \APACyear {2019}}%
}]{%
sun2019deep}
\APACinsertmetastar {%
sun2019deep}%
\begin{APACrefauthors}%
Sun, K.%
, Xiao, B.%
, Liu, D.%
\BCBL {}\ \BBA {} Wang, J.%
\end{APACrefauthors}%
\unskip\
\newblock
\APACrefYearMonthDay{2019}{}{}.
\newblock
{\BBOQ}\APACrefatitle {Deep high-resolution representation learning for human pose estimation} {Deep high-resolution representation learning for human pose estimation}.{\BBCQ}
\newblock
\BIn{} \APACrefbtitle {Proceedings of the IEEE/CVF conference on computer vision and pattern recognition} {Proceedings of the ieee/cvf conference on computer vision and pattern recognition}\ (\BPGS\ 5693--5703).
\PrintBackRefs{\CurrentBib}

\bibitem [\protect \citeauthoryear {%
Tan%
\ \protect \BOthers {.}}{%
Tan%
\ \protect \BOthers {.}}{%
{\protect \APACyear {2025}}%
}]{%
tan2025devbench}
\APACinsertmetastar {%
tan2025devbench}%
\begin{APACrefauthors}%
Tan, A\BPBI W\BPBI M.%
, Yu, S.%
, Long, B.%
, Ma, W\BPBI A.%
, Murray, T.%
, Silverman, R\BPBI D.%
\BDBL {}Frank, M\BPBI C.%
\end{APACrefauthors}%
\unskip\
\newblock
\APACrefYearMonthDay{2025}{{\APACmonth{01}}}{}.
\newblock
{\BBOQ}\APACrefatitle {{{DevBench}}: {{A}} Multimodal Developmental Benchmark for Language Learning} {{{DevBench}}: {{A}} multimodal developmental benchmark for language learning}.{\BBCQ}
\newblock
\BIn{} \APACrefbtitle {Advances in {{Neural Information Processing Systems}}} {Advances in {{Neural Information Processing Systems}}}\ (\BVOL~37, \BPGS\ 77445--77467).
\newblock
\APACaddressPublisher{Vancouver, BC}{}.
\PrintBackRefs{\CurrentBib}

\bibitem [\protect \citeauthoryear {%
Tkachenko%
, Malyuk%
, Holmanyuk%
\BCBL {}\ \BBA {} Liubimov%
}{%
Tkachenko%
\ \protect \BOthers {.}}{%
{\protect \APACyear {2020-2022}}%
}]{%
LabelStudio}
\APACinsertmetastar {%
LabelStudio}%
\begin{APACrefauthors}%
Tkachenko, M.%
, Malyuk, M.%
, Holmanyuk, A.%
\BCBL {}\ \BBA {} Liubimov, N.%
\end{APACrefauthors}%
\unskip\
\newblock
\APACrefYearMonthDay{2020-2022}{}{}.
\newblock
\APACrefbtitle {{Label Studio}: Data labeling software.} {{Label Studio}: Data labeling software.}
\newblock
\begin{APACrefURL} \url{https://github.com/heartexlabs/label-studio} \end{APACrefURL}
\newblock
\APACrefnote{Open source software available from https://github.com/heartexlabs/label-studio}
\PrintBackRefs{\CurrentBib}

\bibitem [\protect \citeauthoryear {%
Tomar%
}{%
Tomar%
}{%
{\protect \APACyear {2006}}%
}]{%
tomar2006converting}
\APACinsertmetastar {%
tomar2006converting}%
\begin{APACrefauthors}%
Tomar, S.%
\end{APACrefauthors}%
\unskip\
\newblock
\APACrefYearMonthDay{2006}{}{}.
\newblock
{\BBOQ}\APACrefatitle {Converting video formats with FFmpeg} {Converting video formats with ffmpeg}.{\BBCQ}
\newblock
\APACjournalVolNumPages{Linux journal}{2006}{146}{10}.
\PrintBackRefs{\CurrentBib}

\bibitem [\protect \citeauthoryear {%
Vong%
, Wang%
, Orhan%
\BCBL {}\ \BBA {} Lake%
}{%
Vong%
\ \protect \BOthers {.}}{%
{\protect \APACyear {2024}}%
}]{%
vong2024}
\APACinsertmetastar {%
vong2024}%
\begin{APACrefauthors}%
Vong, W\BPBI K.%
, Wang, W.%
, Orhan, A\BPBI E.%
\BCBL {}\ \BBA {} Lake, B\BPBI M.%
\end{APACrefauthors}%
\unskip\
\newblock
\APACrefYearMonthDay{2024}{}{}.
\newblock
{\BBOQ}\APACrefatitle {Grounded language acquisition through the eyes and ears of a single child} {Grounded language acquisition through the eyes and ears of a single child}.{\BBCQ}
\newblock
\APACjournalVolNumPages{Science}{383}{6682}{504--511}.
\PrintBackRefs{\CurrentBib}

\bibitem [\protect \citeauthoryear {%
Warstadt%
\ \protect \BOthers {.}}{%
Warstadt%
\ \protect \BOthers {.}}{%
{\protect \APACyear {2023}}%
}]{%
warstadt2023findings}
\APACinsertmetastar {%
warstadt2023findings}%
\begin{APACrefauthors}%
Warstadt, A.%
, Mueller, A.%
, Choshen, L.%
, Wilcox, E.%
, Zhuang, C.%
, Ciro, J.%
\BDBL {}others%
\end{APACrefauthors}%
\unskip\
\newblock
\APACrefYearMonthDay{2023}{}{}.
\newblock
{\BBOQ}\APACrefatitle {Findings of the BabyLM Challenge: Sample-efficient pretraining on developmentally plausible corpora} {Findings of the babylm challenge: Sample-efficient pretraining on developmentally plausible corpora}.{\BBCQ}
\newblock
\BIn{} \APACrefbtitle {Proceedings of the BabyLM Challenge at the 27th Conference on Computational Natural Language Learning.} {Proceedings of the babylm challenge at the 27th conference on computational natural language learning.}
\PrintBackRefs{\CurrentBib}

\bibitem [\protect \citeauthoryear {%
Warstadt%
\ \protect \BOthers {.}}{%
Warstadt%
\ \protect \BOthers {.}}{%
{\protect \APACyear {2020}}%
}]{%
warstadt-etal-2020-blimp-benchmark}
\APACinsertmetastar {%
warstadt-etal-2020-blimp-benchmark}%
\begin{APACrefauthors}%
Warstadt, A.%
, Parrish, A.%
, Liu, H.%
, Mohananey, A.%
, Peng, W.%
, Wang, S\BHBI F.%
\BCBL {}\ \BBA {} Bowman, S\BPBI R.%
\end{APACrefauthors}%
\unskip\
\newblock
\APACrefYearMonthDay{2020}{}{}.
\newblock
{\BBOQ}\APACrefatitle {{BL}i{MP}: The Benchmark of Linguistic Minimal Pairs for {E}nglish} {{BL}i{MP}: The benchmark of linguistic minimal pairs for {E}nglish}.{\BBCQ}
\newblock
\APACjournalVolNumPages{Transactions of the Association for Computational Linguistics}{8}{}{377--392}.
\newblock
\begin{APACrefURL} \url{https://aclanthology.org/2020.tacl-1.25} \end{APACrefURL}
\newblock
\begin{APACrefDOI} \doi{10.1162/tacl_a_00321} \end{APACrefDOI}
\PrintBackRefs{\CurrentBib}

\bibitem [\protect \citeauthoryear {%
Xu%
, Zhang%
, Zhang%
\BCBL {}\ \BBA {} Tao%
}{%
Xu%
\ \protect \BOthers {.}}{%
{\protect \APACyear {2022}}%
}]{%
xu2022vitpose}
\APACinsertmetastar {%
xu2022vitpose}%
\begin{APACrefauthors}%
Xu, Y.%
, Zhang, J.%
, Zhang, Q.%
\BCBL {}\ \BBA {} Tao, D.%
\end{APACrefauthors}%
\unskip\
\newblock
\APACrefYearMonthDay{2022}{}{}.
\newblock
{\BBOQ}\APACrefatitle {Vitpose: Simple vision transformer baselines for human pose estimation} {Vitpose: Simple vision transformer baselines for human pose estimation}.{\BBCQ}
\newblock
\APACjournalVolNumPages{Advances in Neural Information Processing Systems}{35}{}{38571--38584}.
\PrintBackRefs{\CurrentBib}

\bibitem [\protect \citeauthoryear {%
Yamins%
\ \protect \BOthers {.}}{%
Yamins%
\ \protect \BOthers {.}}{%
{\protect \APACyear {2014}}%
}]{%
yamins2014performance}
\APACinsertmetastar {%
yamins2014performance}%
\begin{APACrefauthors}%
Yamins, D\BPBI L.%
, Hong, H.%
, Cadieu, C\BPBI F.%
, Solomon, E\BPBI A.%
, Seibert, D.%
\BCBL {}\ \BBA {} DiCarlo, J\BPBI J.%
\end{APACrefauthors}%
\unskip\
\newblock
\APACrefYearMonthDay{2014}{}{}.
\newblock
{\BBOQ}\APACrefatitle {Performance-optimized hierarchical models predict neural responses in higher visual cortex} {Performance-optimized hierarchical models predict neural responses in higher visual cortex}.{\BBCQ}
\newblock
\APACjournalVolNumPages{Proceedings of the national academy of sciences}{111}{23}{8619--8624}.
\PrintBackRefs{\CurrentBib}

\bibitem [\protect \citeauthoryear {%
Yoshida%
\ \BBA {} Smith%
}{%
Yoshida%
\ \BBA {} Smith%
}{%
{\protect \APACyear {2008}}%
}]{%
yoshida2008}
\APACinsertmetastar {%
yoshida2008}%
\begin{APACrefauthors}%
Yoshida, H.%
\BCBT {}\ \BBA {} Smith, L\BPBI B.%
\end{APACrefauthors}%
\unskip\
\newblock
\APACrefYearMonthDay{2008}{}{}.
\newblock
{\BBOQ}\APACrefatitle {What's in view for toddlers? Using a head camera to study visual experience} {What's in view for toddlers? using a head camera to study visual experience}.{\BBCQ}
\newblock
\APACjournalVolNumPages{Infancy}{13}{3}{229--248}.
\PrintBackRefs{\CurrentBib}

\bibitem [\protect \citeauthoryear {%
Yu%
, Zhang%
, Slone%
\BCBL {}\ \BBA {} Smith%
}{%
Yu%
\ \protect \BOthers {.}}{%
{\protect \APACyear {2021}}%
}]{%
yu2021infant}
\APACinsertmetastar {%
yu2021infant}%
\begin{APACrefauthors}%
Yu, C.%
, Zhang, Y.%
, Slone, L\BPBI K.%
\BCBL {}\ \BBA {} Smith, L\BPBI B.%
\end{APACrefauthors}%
\unskip\
\newblock
\APACrefYearMonthDay{2021}{}{}.
\newblock
{\BBOQ}\APACrefatitle {The infant’s view redefines the problem of referential uncertainty in early word learning} {The infant’s view redefines the problem of referential uncertainty in early word learning}.{\BBCQ}
\newblock
\APACjournalVolNumPages{Proceedings of the National Academy of Sciences}{118}{52}{e2107019118}.
\PrintBackRefs{\CurrentBib}

\bibitem [\protect \citeauthoryear {%
YUAN%
\ \protect \BOthers {.}}{%
YUAN%
\ \protect \BOthers {.}}{%
{\protect \APACyear {2021}}%
}]{%
NEURIPS2021_3bbfdde8}
\APACinsertmetastar {%
NEURIPS2021_3bbfdde8}%
\begin{APACrefauthors}%
YUAN, Y.%
, Fu, R.%
, Huang, L.%
, Lin, W.%
, Zhang, C.%
, Chen, X.%
\BCBL {}\ \BBA {} Wang, J.%
\end{APACrefauthors}%
\unskip\
\newblock
\APACrefYearMonthDay{2021}{}{}.
\newblock
{\BBOQ}\APACrefatitle {HRFormer: High-Resolution Vision Transformer for Dense Predict} {Hrformer: High-resolution vision transformer for dense predict}.{\BBCQ}
\newblock
\BIn{} M.~Ranzato, A.~Beygelzimer, Y.~Dauphin, P.~Liang\BCBL {}\ \BBA {} J\BPBI W.~Vaughan\ (\BEDS), \APACrefbtitle {Advances in Neural Information Processing Systems} {Advances in neural information processing systems}\ (\BVOL~34, \BPGS\ 7281--7293).
\newblock
\APACaddressPublisher{}{Curran Associates, Inc.}
\newblock
\begin{APACrefURL} \url{https://proceedings.neurips.cc/paper_files/paper/2021/file/3bbfdde8842a5c44a0323518eec97cbe-Paper.pdf} \end{APACrefURL}
\PrintBackRefs{\CurrentBib}

\bibitem [\protect \citeauthoryear {%
Zhuang%
, Fedorenko%
\BCBL {}\ \BBA {} Andreas%
}{%
Zhuang%
\ \protect \BOthers {.}}{%
{\protect \APACyear {2023}}%
}]{%
zhuang2023visual}
\APACinsertmetastar {%
zhuang2023visual}%
\begin{APACrefauthors}%
Zhuang, C.%
, Fedorenko, E.%
\BCBL {}\ \BBA {} Andreas, J.%
\end{APACrefauthors}%
\unskip\
\newblock
\APACrefYearMonthDay{2023}{}{}.
\newblock
{\BBOQ}\APACrefatitle {Visual Grounding Helps Learn Word Meanings in Low-Data Regimes} {Visual grounding helps learn word meanings in low-data regimes}.{\BBCQ}
\newblock
\APACjournalVolNumPages{arXiv preprint arXiv:2310.13257}{}{}{}.
\PrintBackRefs{\CurrentBib}

\bibitem [\protect \citeauthoryear {%
Zhuang%
, Fedorenko%
\BCBL {}\ \BBA {} Andreas%
}{%
Zhuang%
\ \protect \BOthers {.}}{%
{\protect \APACyear {2024}}%
}]{%
zhuang2024lexicon}
\APACinsertmetastar {%
zhuang2024lexicon}%
\begin{APACrefauthors}%
Zhuang, C.%
, Fedorenko, E.%
\BCBL {}\ \BBA {} Andreas, J.%
\end{APACrefauthors}%
\unskip\
\newblock
\APACrefYearMonthDay{2024}{}{}.
\newblock
{\BBOQ}\APACrefatitle {Lexicon-Level Contrastive Visual-Grounding Improves Language Modeling} {Lexicon-level contrastive visual-grounding improves language modeling}.{\BBCQ}
\newblock
\APACjournalVolNumPages{arXiv preprint arXiv:2403.14551}{}{}{}.
\PrintBackRefs{\CurrentBib}

\bibitem [\protect \citeauthoryear {%
Zhuang%
, She%
, Andonian%
, Mark%
\BCBL {}\ \BBA {} Yamins%
}{%
Zhuang%
\ \protect \BOthers {.}}{%
{\protect \APACyear {2020}}%
}]{%
zhuang2020unsupervised}
\APACinsertmetastar {%
zhuang2020unsupervised}%
\begin{APACrefauthors}%
Zhuang, C.%
, She, T.%
, Andonian, A.%
, Mark, M\BPBI S.%
\BCBL {}\ \BBA {} Yamins, D.%
\end{APACrefauthors}%
\unskip\
\newblock
\APACrefYearMonthDay{2020}{}{}.
\newblock
{\BBOQ}\APACrefatitle {Unsupervised learning from video with deep neural embeddings} {Unsupervised learning from video with deep neural embeddings}.{\BBCQ}
\newblock
\BIn{} \APACrefbtitle {Proceedings of the ieee/cvf conference on computer vision and pattern recognition} {Proceedings of the ieee/cvf conference on computer vision and pattern recognition}\ (\BPGS\ 9563--9572).
\PrintBackRefs{\CurrentBib}

\bibitem [\protect \citeauthoryear {%
Zhuang%
\ \protect \BOthers {.}}{%
Zhuang%
\ \protect \BOthers {.}}{%
{\protect \APACyear {2022}}%
}]{%
zhuang2022well}
\APACinsertmetastar {%
zhuang2022well}%
\begin{APACrefauthors}%
Zhuang, C.%
, Xiang, Z.%
, Bai, Y.%
, Jia, X.%
, Turk-Browne, N.%
, Norman, K.%
\BDBL {}Yamins, D.%
\end{APACrefauthors}%
\unskip\
\newblock
\APACrefYearMonthDay{2022}{}{}.
\newblock
{\BBOQ}\APACrefatitle {How Well Do Unsupervised Learning Algorithms Model Human Real-time and Life-long Learning?} {How well do unsupervised learning algorithms model human real-time and life-long learning?}{\BBCQ}
\newblock
\APACjournalVolNumPages{Advances in Neural Information Processing Systems}{35}{}{22628--22642}.
\PrintBackRefs{\CurrentBib}

\bibitem [\protect \citeauthoryear {%
Zhuang%
\ \protect \BOthers {.}}{%
Zhuang%
\ \protect \BOthers {.}}{%
{\protect \APACyear {2021}}%
}]{%
zhuang2021}
\APACinsertmetastar {%
zhuang2021}%
\begin{APACrefauthors}%
Zhuang, C.%
, Yan, S.%
, Nayebi, A.%
, Schrimpf, M.%
, Frank, M\BPBI C.%
, DiCarlo, J\BPBI J.%
\BCBL {}\ \BBA {} Yamins, D\BPBI L.%
\end{APACrefauthors}%
\unskip\
\newblock
\APACrefYearMonthDay{2021}{}{}.
\newblock
{\BBOQ}\APACrefatitle {Unsupervised neural network models of the ventral visual stream} {Unsupervised neural network models of the ventral visual stream}.{\BBCQ}
\newblock
\APACjournalVolNumPages{Proceedings of the National Academy of Sciences}{118}{3}{e2014196118}.
\PrintBackRefs{\CurrentBib}

\end{thebibliography}
\bibliographystyle{ccn_style}

\clearpage
\appendix
\section{Appendix}

\subsection{Dataset details}

\subsubsection{Participant consent} All data collection was approved under
Stanford University Protocols \#20398 and \#72325. 
consent was obtained via one-on-one conversations. Given the sensitive nature of the data, families had multiple opportunities to withdraw their recordings. They could mark videos for deletion during recording and up to six months during the embargo period. 

\subsubsection{Participant instructions \& recording details}
All participant instructions were taken from \cite{long2023} which developed the protocols for using the BabyView Camera, and are publicly available at https://osf.io/kwvxu/.

Families were instructed to record as often as was feasible for their families, with a requested minimum of 45 minutes per week. We used standard, rechargeable 9V battery to provide power to the BabyView camera, which allows for continuous 45-60 minute recordings on a standard charge. Families were compensated based on the duration (mins) of video recordings they provided on a weekly basis, as well as bonuses for questionnaires, totalling 18,370.00 USD across all families.

\subsubsection{BV-Home additional participant demographics}

Our sample is highly educated, with 24/31 families having at least one parent with a graduate degree, and with all but one family having at least one parent with a 4-year college degree. 14/31 children are exposed to more than one language at home, including the following languages: English, Chinese, Farsi, French, Gujarati, Japanese, Korean, Malayalam, Portuguese, Spanish, Tagalog, Thai, Vietnamese. Geographically, 22/31 of families live within California, 5/31 live in the Northeastern United States, 1/31 live in the Southern United States, 1/31 live in the Midwestern United States, 1/31 live in Canada, and 1/31 live in South Korea.

Participating children were 58.06\% female, 41.94\% male, 0\% African American/Black, 19.35\% Asian American/Pacific Islander, 38.71\% Caucasian/White, 9.68\% Hispanic/Latinx, 41.94\% multiracial, 0\% other.

We only have income information for 27/31 families, as reporting was optional. The average family income of our sample is 231622.15 USD (75000–1000000 USD, SD = 181466.57 USD). 16/31 families have more than one child in the household and 2/31 families have more than 2 caregivers living in the household.

\subsubsection{BV-Home language outcome questionnaires}
 Long-form MacArthur Bates CDI language questionnaires (https://mb-cdi.stanford.edu/) were administered every 3 months starting at enrollment. Families were provided compensation for each completed questionnaire.  These parent-report forms assess children's vocabulary comprehension and production; aggregate data by age can be viewed at wordbank.stanford.edu. Forms were administered through Web-CDI (https://webcdi.org/). Data from 69 (4 Spanish, 65 English) questionnaires are included in this first release of the dataset.

\subsubsection{Video processing pipeline}
Videos were manually uploaded by each family to their personalized Google Drive folders. The uploaded videos were automatically downloaded to a secure server where the metadata (accelerometer and gyroscope) were extracted and the videos were compressed then uploaded to a second Google Drive platform. The compression step used the ffmpeg \citep{tomar2006converting} program to encode video into the libx265 format with a constant rate factor of 23 to enable high quality MP4 videos.

\subsection{Annotation details}

\subsubsection{Pose keypoint details and evaluation}
 The pose keypoints that were evaluated includes 17 keypoints: nose, left eye, right eye, left ear, right ear, left shoulder, right shoulder, left elbow, right elbow, left wrist, right wrist, left hip, right hip, left knee, right knee, left ankle, and right ankle.

 The Object Keypoint Similarity (OKS) metric reported is as follows:
 
\[
OKS = \frac{\sum_i \exp\left(-\frac{d_i^2}{2s^2k_i^2}\right)\delta(v_i>0)}{\sum_i \delta(v_i>0)}.
\]
In this formula, \(d_i\) represents the Euclidean distance between the detected keypoint and the ground truth, \(v_i\) indicates the visibility of the ground truth keypoint, \(s\) denotes the object scale, and \(k_i\) is a constant specific to each keypoint that adjusts the falloff. We report standard metrics for average precision and recall:  AP (the average of AP scores at 10 different OKS thresholds: 0.50, 0.55, ..., 0.90, 0.95), and AR (the average of AR scores at OKS = 0.50, 0.55, ..., 0.90, 0.95).

\subsubsection{Compute resources and infrastructure for annotation}
Our annotation work was performed on an internal cluster server with an AMD EPYC 9334 32-Core Processor, 756GB memory, 8 NVIDIA A40 GPUs, and Ubuntu 20.04. We used 8 GPUs for speech recognition and 1 GPU for both assisting with annotation and testing pose detection models on the validation set.

\subsection{Language benchmark details}
\subsubsection{Language model training \& evaluation details}

% We chose a small causal language model (LM) such as GPT-2 as an initial measurement of the ability of these smaller datasets to act as a good learning signal for language in LMs. This is partially inspired by the BabyLM challenge [CITE], which aims to mimic the data limitations of human development by training LMs using similar amounts of linguistic data available to a child while growing up. This is magnitudes less than the data used to pretrain present-day LLMs. BabyLM mainly assesses two main capabilities of models: 1) grammar (using BLiMP [CITE]) and 2) language understanding (using SuperGLUE [CITE]).

In training our GPT-2 models, we used a learning rate (LR) of $10^{-4}$, linear LR scheduler with no warmup steps, a batch size of 8 or 16 per GPU, training seeds of 0, 42, and 123, and Adam optimizer with $\beta=(0.9,0.999)$ and $\epsilon=10^{-8}$.

The final chosen GPT-2 model for each experiment is the epoch that performed best (lowest loss) on the corresponding validation split. \textcolor{black}{The corresponding byte-level BPE tokenizer for each model was also trained from scratch on the corresponding dataset.} 

The training data was set up so that each line corresponded to a single transcribed conversation, which is broken up into chunks of 1024 consecutive tokens by GPT-2 during training. To ensure the data format is consistent for evaluation purposes, we aligned the most important and frequently occurring speaker labels across datasets (mainly based on the existing CHILDES labels): CHI for the target child, MOT for the mother or female adult, and OCHI for other children. All other speaker labels were kept to their default. Around 60\% or more of all utterances within each dataset were from CHI or MOT.

See below for an example of part of a single training conversation. Double asterisks surround speaker labels, double newline tokens separate utterances, and an end-of-text token marks the end of the conversation. This format was consistent across all conversations and datasets.

\textit{**CHI**: Hi. \escape{n}\escape{n} **CHI**: There you go. \escape{n}\escape{n} **OCHI**: Do you have a little ball in your cup. \escape{n}\escape{n} (...) \escape{n}\escape{n} **CHI**: Are those your stars? \escape{n}\escape{n} **MOT**: Can you say star? \escape{n}\escape{n} **CHI**: Star. \escape{n}\escape{n} **CHI**: Look. \escape{n}\escape{n} **CHI**: Stars. \escape{n}\escape{n} **MOT**: Stars. See? Look, look at the yellow star, a golden star. \texttt{<|endoftext|>}}

We found cases of duplicate conversations and duplicate utterances within conversations among the transcribed data across the three datasets. We removed these to the best of our ability before training. 

The Zorro evaluation was inspired by BLiMP \citep{warstadt-etal-2020-blimp-benchmark} and is a modification for child-directed language (e.g. lower vocabulary). However, it was designed specifically for masked language models such as RoBERTa. To adapt it to GPT-2, we reformatted the Zorro data to match the BLiMP format and used the BLiMP evaluation in the BabyLM evaluation suite \footnote{\url{https://github.com/babylm/evaluation-pipeline-2023}} since the main difference between the two is the evaluation data. Further, we use the full Zorro test suite and do not filter examples by vocabulary. Hence, our results are not comparable to \citet{qin2024systematic} which filters Zorro examples by the vocabulary of their training datasets.

To better match the training data format and assess the effects of speaker labels on evaluation, we came up with three variations of Zorro: 1) the original Zorro evaluation sentences, 2) the sentences with the CHI speaker label prepended, and 3) the sentences with the MOT speaker label prepended. To further match the training data, the speaker labels were surrounded by double asterisks, and sentences included double newline tokens (before and after). We found that variation 3 (prepending each Zorro sentence with MOT) worked best for all datasets. This is likely because the utterances spoken by the mother and female adults are typically more grammatical than those of the child.

%As seen in Table \ref{tab:zorro_results}, all models perform better when the evaluation data is more closely aligned with the training data format (2nd or 3rd variation of Zorro sentences), especially with the MOT speaker label (3rd variation). 

\begin{comment}
\subsubsection{Detailed Language Model Experiment Results}

See Table \ref{tab:zorro_results} for the Zorro evaluation results of our GPT-2 models, along with the best Zorro evaluation format for each; Zorro results indicate the final average values.

\begin{table}[ht!]
\caption{Quantitative results on the Zorro benchmark}
\label{tab:zorro_results}
\centering
\begin{tabular}{lcc}
\toprule
% & \multicolumn{2}{c}{ImageNet} & NYUv2 - Depth & COCOStuff - Segmentation \\
\textbf{Model} & \textbf{Zorro} & \textbf{Best Eval. Format} \\
\midrule
BV-Home                     & 64.13\% & CHI\\
SAYCam                      & 64.06\% & MOT\\
CHILDES (2M)                & 66.57\% & MOT\\
SAYCam + BV-Home            & 69.39\% & CHI\\
CHILDES (4M)                & 69.76\% & MOT\\
CHILDES (20M)               & 77.77\% & MOT\\
\bottomrule
\end{tabular}
\end{table}
\end{comment}

\subsubsection{Compute resources and infrastructure for language model training}

Our language model experiments were run on an internal cluster server consisting of 5 A40s and 8 A100s.%(80GB VRAM each).%, and a cloud provider VM instance consisting of four A100s (80GB VRAM each).

\subsection{Vision benchmark details}

\subsubsection{Video preprocessing}

\paragraph{BabyView} We sample BV-Home at 5 FPS at a resolution of 720x360 for the initial 224 global crop training of DINO, and at 720x1280 for the 518 high resolution final stage of training. This results in a total of 16M frames for the total BV-Home dataset. 

To create datasets of different sizes (1\%, 5\%, etc.) we randomly selected complete clips and append them to a continuously increasing list which we save at different size increments. This ensureed that every smaller set of data was a strict subset of the larger set (e.g., the clips in the 1\% set are all contained in 5\% set etc.). After getting these lists of clips, we extracted frames with the same procedure.

 \textcolor{black}{Because the dataset was at a 9:16 widescreen aspect ratio, significantly different from the mostly 4:3 ImageNet image aspect ratio for which the DINO random cropping strategy was developed, we took random crop with aspect ratio in the 4:3 to 3:4 range with the biggest possible size, before performing the DINO cropping and augmentation. Empirically this resulted in a 1\% improvement in ImageNet classification accuracy.}

\paragraph{SAYCam} We sampled SAYCam at 5 FPS in the native resolution of 480x640. This resulted in a total of 8.5M frames.

\paragraph{Ego4D} We took the complete Ego4D dataset without additional post-processing and sample frames at 1 FPS using ffmpeg at 1/2 of the original resolution. The smallest side of the images we extracted ranged from 360 to 960 pixels---sufficient resolution for training (the variance in resolution exists in the original dataset due to the use of different recording devices). We reduced the original resolution to limit the footprint of the dataset on disk and to lower the computational cost of data loading. This resulted in a total of 15M frames. We apply the same 3:4 aspect ratio augmentation that we did for BabyView.

\subsubsection{Training} 
\paragraph{DINOv2} To train DINOv2 we use the official code repository.\footnote{\url{https://github.com/facebookresearch/dinov2}} We tried to perform minimal modifications of the existing pipeline. We trained a ViT-B/14 with a batch size of 1024 with the default ImageNet-1K training config for the default 125K parameter updates. This initial training is done with a global crop of 224x224. All other hyperparameters were kept the same. We experimented with doubling the amount of parameter updates but did not see improvements.  Following the DINOv2 paper, we trained for an additional 10K parameter updates with a global crop of size 518x518.

\paragraph{MoCov3} To train MoCov3 we use the official code repository.\footnote{\url{https://github.com/facebookresearch/moco-v3}} We trained a ViT-B/16 with a batch size of 512 with the default ImageNet-1K training configurations for up to 725K parameter updates. Similar to DINOv2, the training was done with an initial global crop of 224x224. 

\subsubsection{Downstream tasks}
\paragraph{ImageNet Category Recognition} We used the code from the official DINOv2 repository for kNN classification or for training a linear classifier. Our evaluation procedure, therefore, directly followed the procedure used in DINOv2.

\paragraph{NYUv2 Depth Estimation} Following the descriptions in the DINOv2 paper, we used the Monocular Depth Toolbox~\citep{lidepthtoolbox2022}, and follow their evaluation protocol. The code interfacing DINOv2 with this package is not released, but the trained depth estimation models and configs are released.  \textcolor{black}{After writing the interface code, we verified that the evaluation was correct by training a DPT-based~\cite{ranftl2021vision} depth estimator (Dense Prediction Transformer) which uses the trained model as a backbone for the semantic segmentation and depth estimation task.} We utilized this codebase on top of an off-of-the shelf official DINOv2 checkpoint which matched the performance from the paper.

\paragraph{COCOStuff Semantic Segmentation} We interfaced the official DINOv2 code with the mmsegmentation package~\citep{mmseg2020}. Similarly, the interface code is not released but the models and configs are available. To verify correctness, we trained a linear probe on top of an off-the-shelf official DINOv2 checkpoint and matched the performance from the paper on PASCAL VOC. We used the same config to train a linear probe on COCOStuff as was released for PASCAL VOC. We did not find improvements by training for longer. Future work may investigate training more complex architectures, which was prohibitive for this work due to the time and compute constraints required.

\subsubsection{Compute resources}
The DINOv2 vision models in this paper can be trained on a single 8x NVIDIA A40 GPU node. While no multi-node training is required, one full training run of DINOv2 takes about 3 days on 8x A40 GPUs. This translates to about 550 GPU hours per experiment, making it difficult to perform multiple runs to obtain error bars.

\subsection{Data accessibility}
No data are available for review due to the parental embargo policy. All data will be hosted on https://nyu.databrary.org/ in June 2025 after the parental embargo for this release has lapsed. Researchers must be affiliated with a PI at a research institution, who must request access to the project. 

All compressed videos and their associated meta-data will be named according to a standardized format that encodes the participant id and the date at which the recordings were made. A .csv spreadsheet will provide detailed, anonymized information about each individual participant. Separate language outcome data (in standard CDI format) will be provided and linked to the individual participant IDs.

\subsection{Licensing}
The code and behavioral data published with the benchmark will be licensed under CC BY-NC 4.0. The video dataset is licensed under the terms laid out in the Databrary Access Agreement, see https://databrary.org/about/agreement/agreement.html.

Licenses for models used: YOLOXPose is licensed under the GPL-3.0 license. MMPose, RTMO, SimCC, ViTPose, mmsegmentation, DINOv2, Monocular Depth Toolbox, and LabelStudio are licensed under the Apache-2.0 license. GPT-2 is licensed under the modified MIT License. RTMPose is licensed under the MIT license. All are permissive for this paper release.

We, the authors, bear all responsibility in case we have violated any rights by the publication of these data and code in these venues. 

\subsection{Code availability}
Relevant model training code can be found at https://osf.io/j45qa/.

\end{document}